\documentclass[letterpaper]{article} 
\usepackage[]{aaai2026}  
\usepackage{times}  
\usepackage{helvet}  
\usepackage{courier}  
\usepackage[hyphens]{url}  
\usepackage{graphicx} 
\usepackage{natbib}  
\usepackage{caption} 
\usepackage{algorithm}
\usepackage{algorithmic}
\usepackage{amsmath}
\usepackage{booktabs}
\usepackage{longtable}
\usepackage{multirow}
\usepackage{tcolorbox}
\usepackage{url}
\tcbuselibrary{breakable}
\usepackage{newfloat}
\usepackage{listings}
\usepackage{makecell}
\DeclareCaptionStyle{ruled}{labelfont=normalfont,labelsep=colon,strut=off}
\floatstyle{ruled}
\newfloat{listing}{tb}{lst}{}
\floatname{listing}{Listing}
\title{Towards Robust and Fair Next Visit Diagnosis Prediction \\ under Noisy Clinical Notes with Large Language Models}
\author {
    Heejoon Koo
}
\affiliations{
    University College London\\
    heejoon.koo.17@alumni.ucl.ac.uk
}

\begin{document}

\maketitle

\begin{abstract}
A decade of rapid advances in artificial intelligence (AI) has opened new opportunities for clinical decision support systems (CDSS), with large language models (LLMs) demonstrating strong reasoning abilities on timely medical tasks. However, clinical texts are often degraded by human errors or failures in automated pipelines, raising concerns about the reliability and fairness of AI-assisted decision-making. Yet the impact of such degradations remains under-investigated, particularly regarding how noise-induced shifts can heighten predictive uncertainty and unevenly affect demographic subgroups. We present a systematic study of state-of-the-art LLMs under diverse text corruption scenarios, focusing on robustness and equity in next-visit diagnosis prediction. To address the challenge posed by the large diagnostic label space, we introduce a clinically grounded label-reduction scheme and a hierarchical chain-of-thought (CoT) strategy that emulates clinicians’ reasoning. Our approach improves robustness and reduces subgroup instability under degraded inputs, advancing the reliable use of LLMs in CDSS. We release code at \url{https://github.com/heejkoo9/NECHOv3}.
\end{abstract}

\section{Introduction}
\label{sec:introduction}

CDSS assist clinicians in making evidence-based diagnostic, prognostic, and therapeutic decisions, improving patient care quality \cite{sutton2020overview}. They have shown significant potential to reduce diagnostic errors, improve treatment adherence, and streamline clinical workflows. In particular, progress in language technologies has improved the translation of unstructured documents into actionable recommendations, narrowing the gap between complex clinical data and point-of-care decision-making \cite{chen2023algorithmic}.

Despite advances, clinical text is often degraded by human error, workflow inefficiencies, and operational constraints \cite{berndt2015case, koo2023comprehensive, koo2023survey, koo2025overcoming}. These degradations take two forms: 1) missingness involving absent entities or laboratory values from time-pressured documentation \cite{kalisch2009missed}, and 2) perturbations introduced during processing. In specific, Automatic Speech Recognition (ASR) can induce homophone substitutions (e.g., \textit{ileum} $\rightarrow$ \textit{ilium}) \cite{latif2020speech}, and Optical Character Recognition (OCR) can scramble characters \cite{biondich2002modern}. While certain forms of textual noise can function as an implicit data augmentation \cite{bayer2022survey}, it may have detrimental effects in high-stakes domains such as healthcare, where minor textual distortions may alter clinically relevant cues, propagate misinformation, or ultimately degrade the reliability and safety of downstream decisions.

Parallel to concerns over robustness, fairness has been a priority in AI and healthcare \cite{chen2023algorithmic}. Ensuring equitable performance across demographic subgroups, such as race, age, and other socio-economic statuses, is essential for trustworthy AI deployment \cite{rajkomar2018ensuring}. Prior works have demonstrated that AI systems often exhibit biases, exacerbating health disparities \cite{chen2023algorithmic}. Moreover, \cite{bhatt2021uncertainty} argue that disparate uncertainty compounds existing unfairness, disproportionately affecting marginalised subgroups.

Recently, LLMs have transformed modern AI with exceptional language understanding, positioning them as promising tools for CDSS \cite{bedi2024testing}. Prior studies indicate that LLMs exhibit notable robustness to textual corruptions \cite{singh2024robustness}, while extensive research has also highlighted their inherent biases and fairness challenges across demographic and linguistic dimensions \cite{li2023survey}.

However, the extent to which textual degradations jointly affect both robustness and fairness in LLM-based CDSS remains under-explored. This gap is particularly consequential in high-stakes healthcare, where clinical narratives often contain domain-specific terminology, irregular syntax, and densely packed information. Such characteristics make both accurate prediction and equitable model performance especially challenging under noisy or imperfect documentation conditions and act as a barrier towards reliable deployment.

To systematically quantify these challenges, we conduct a comprehensive evaluation of state-of-the-art LLMs on the publicly available MIMIC-IV dataset \cite{johnson2023mimic}. We simulate realistic documentation artifacts commonly encountered in clinical environments, including lab-value omission and prior-note duplication, as well as perturbations inspired by ASR-induced homophone substitutions and OCR-style visual confusions. These corruptions serve to test model reliability while enabling the assessment of fairness disparities across demographic subgroups under degraded input conditions.

Our primary focus is on next-visit diagnosis prediction, a foundational task for anticipatory care in intensive care units (ICUs) \cite{koo2024next, koo2025overcoming}. A key challenge in prior work on LLM-driven diagnosis prediction is the large and heterogeneous label space, which undermines zero-shot performance. To address this, we introduce a novel approach, \textbf{NECHO v3}, that integrates 1) a label-reduction mapping informed by clinical ontologies, and 2) Chain-of-Thought (CoT) guided hierarchical reasoning, explicitly mirroring the diagnostic reasoning process of clinicians. This design improves both robustness and fairness, providing a structured framework for LLM-based clinical prediction in real-world settings.

\section{Related Works}
\label{sec:related_works}

\subsection{AI for Healthcare: from ML to LLM}
\label{sec:related_works_ml_llm_healthcare}

Clinical predictive models have progressed from feature-engineered models on structured EHRs for risk stratification to longitudinal diagnosis prediction models \cite{ koo2024next}. This has significantly improved accuracy but requires task-specific modelling \cite{rajkomar2018ensuring}. Recently, LLMs shift the paradigm towards general-purpose reasoning over clinical text, enabling zero-/few-shot prompting, instruction tuning, and tool integration; applications span summarisation, triage, clinical QA, alongside ethical and governance considerations \cite{bedi2024testing}.

\subsection{Text Degradation in Clinical Settings}
\label{sec:related_works_text_degradation}

Text degradation, manifested as both missingness and perturbation, is pervasive in clinical documentation and may, in some cases, incidentally serve as a form of data augmentation by increasing textual variability \cite{shorten2021text}. However, in high-stakes clinical contexts, such degradations typically undermine model reliability, interpretability, and ultimately the safety of decision-making processes \cite{moradi2021deep}.

Consequently, previous work has proposed mitigation strategies \cite{getzen2023mining} and evaluated the impact of noise on clinical text modelling tasks \cite{wu2025pay}. Recent studies have begun to probe LLMs' robustness under textual corruption \cite{singh2024robustness}; however, most analyses lack 1) reflecting plausible corruption scenarios in the real-world, and 2) evaluation across subgroups.

\subsection{Fairness in Healthcare}
\label{sec:related_works_fairness_healthcare}

AI models inherit and amplify societal biases embedded in their training data, which can lead to uneven predictive performance across demographic sub-populations \cite{li2023survey}. In healthcare, these disparities trigger biased prediction risks, reinforcing existing inequities in diagnosis, treatment, and resource allocation \cite{rajkomar2018ensuring, zhao2024can}. 

To address these concerns, the NLP community has developed benchmarks to quantify bias in LLMs, targeting tasks such as commonsense reasoning \cite{shen2024generalization} and evaluation in healthcare \cite{zhang2024climb}. However, there remains an absence of studies exploring how text degradation affects fairness in medical prediction for subgroup populations, underscoring the need for fairness assessments under noisy textual inputs.

\paragraph{Positioning.} The most closely related work to ours is \cite{zhao2024can}, which investigates fairness in diagnosis prediction using LLMs. We extend this line of research by: 1) evaluating longitudinal multi-label next visit diagnosis on the publicly available MIMIC-IV dataset; 2) introducing a clinically grounded label-reduction and hierarchical CoT prompting for diagnosis prediction; 3) simulating corruption processes in realistic clinical settings; and 4) systematically assessing robustness and fairness across demographic subgroups. Thus, this paper enables a more comprehensive evaluation of LLMs for real-world CDSS.

\section{Methodologies}
\label{sec:methodologies}

\subsection{Problem Formulation}
\label{sec:problem_formulation}

\textbf{Next Visit Diagnosis Prediction.} A patient \( p \) has clinical records, \( V_1, \ldots, V_T \), where each visit \( V_t \) contains a clinical note \( N_t \). The objective is to predict the multiple diagnoses \( C_{T+1} \) that will appear in the next visit \( V_{T+1} \), based upon the patient’s longitudinal clinical data.

\begin{figure*}[!ht]
    \centering
    \includegraphics[width=0.8\textwidth]{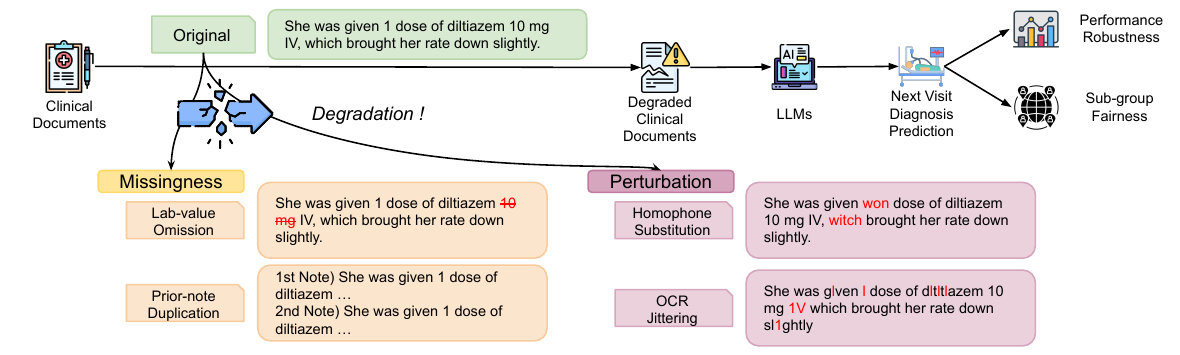}
    \caption{An Overview of Our LLM Evaluation Pipeline under Various Clinical Note Degradation.}
    \label{fig:llm_eval_framework}
\end{figure*}

\subsection{Prompt}
\label{sec:prompt}

We employ a zero-shot CoT prompting strategy \cite{wei2022chain} designed to imitate clinicians’ reasoning for next visit diagnosis prediction. LLM is first prompted to hypothesise multiple plausible parental level diagnostic categories, then enumerate the most likely target diagnoses within those categories, inspired by \cite{koo2024next}. Then, it is provided with the complete set of clinical notes of a patient along with basic demographic information. Full prompts are provided in Appendix \ref{sec:appendix_prompt}.

\subsection{Text Degradation}
\label{sec:text_degradation}

In real-world clinical documentation, textual degradation arises from many uncertain factors \cite{koo2025overcoming}. We organise degradations into two families of missingness and perturbation to enable systematic evaluation for model robustness. Missingness captures lab-value omissions as well as note-level copy-forward artifacts. Perturbations stem from system-induced noise from ASR (e.g., homophone substitutions) and OCR (e.g., character jittering). Although more text degradation scenarios are possible, we restrict our analysis to the defined scenarios.

\textbf{Lab-value Omission} removes numeric laboratory results (e.g., blood glucose, creatinine) from the note, reflecting masking due to privacy or omission in timely settings \cite{kalisch2009missed}.

\textbf{Prior-note Duplication} inserts a verbatim duplicate of a prior note into the current note to model copy-forward artifacts \cite{o2009physicians}.

\textbf{Homophone Substitution} replaces words with phonetically similar alternatives to mimic ASR mis-recognitions (e.g., pain → pane) \cite{latif2020speech}.

\textbf{OCR Jittering} introduces character-level confusions typical of OCR (e.g., history → h1story; O$ \leftrightarrow $ 0, l$ \leftrightarrow $ 1, rn $ \leftrightarrow $ m) \cite{biondich2002modern}.

\section{Experiments}
\label{sec:experiments}

\subsection{Dataset and Preprocessing}
\label{sec:dataset_preprocessing}

\paragraph{Dataset.}
We conduct experiments on the MIMIC-IV dataset \cite{johnson2023mimic}, a large-scale, de-identified EHR repository from Beth Israel Deaconess Medical Center. The corpus comprises structured data (e.g., demographics, diagnosis codes) and unstructured clinical notes authored by healthcare providers. Although the dataset may inherently contain noise, such as inconsistent abbreviations, irregular formatting, and typographical errors, we assume that the data is largely clean. 

\paragraph{Dataset Pre-processing.}
We largely follow prior work on next visit diagnosis prediction \cite{koo2024next, koo2025overcoming}. We stratify the patient population by age into three groups: 18–40, 41–60, 61+, and by race into six categories: Asian, Black, Hispanic/Latino, Other, Unknown, and White. Other group is for those whose racial identification falls outside pre-defined categories and the Unknown group is for those whose racial information is missing or not obtained.

For clinical notes, we use only discharge summaries and adopt the pre-processing pipeline by Clinical Longformer \cite{li2022clinical}: removing de-identification placeholders, normalising punctuation and non-alphanumeric symbols, lowercasing, and trimming whitespace. We further remove sex information for fairness quantification.

For diagnosis codes, we map ICD-9-CM diagnoses to the HCUP Clinical Classifications Software (CCS) hierarchy: 17 multi-level diagnostic chapters (multi-level CCS\footnote{https://hcup-us.ahrq.gov/toolssoftware/ccs/AppendixCMultiDX.txt}) and 295 single-level categories (single-level CCS\footnote{https://hcup-us.ahrq.gov/toolssoftware/ccs/AppendixASingleDX.txt}). To reduce dimensionality while retaining clinical validity, we merge sparse or overlapping categories with the aid of ICD ontology, thereby yielding 17 parental-level diagnostic systems and 46 child-level clinical sub-categories. Each patient receives an average of 9 diagnoses, with a maximum of 25. We provide this mapping in Appendix \ref{sec:appendix_diagnosis_mapping}.

Finally, we restrict the cohort to patients with at least two visits and consider only their most recent five visits. Also, we do not include records with non-positive lengths of stay.

\paragraph{Text Degradation Simulation.}
We simulate text corruption by: lab-value omission (20–30\%), prior-note duplication (30–40\%), homophone substitution (5–15\%), and OCR jittering (5–15\%). We apply a single degradation per note and do not stack multiple degradations.

\paragraph{Sub-population Stratification.}
We sample 8k patients, balancing API cost while maintaining subgroup distributions. Summary statistics are reported in Table~\ref{tab:demographics_stat}.

\begin{table}[ht]
\centering
\scriptsize
\begin{tabular}{cccc}
\toprule
\midrule
\textbf{Variable} & \textbf{Category} & \textbf{Count} & \textbf{Proportion (\%)} \\
\hline
\midrule
\multirow{6}{*}{Race}
 & White              &   5,742 & 71.78 \\
 & Black              &   1,153 & 14.41 \\
 & Hispanic/Latino    &   381 & 4.76 \\
 & Other              &   263 & 3.29 \\
 & Asian              &   240 & 3.00 \\
 & Unknown            &   221 & 2.76 \\
\midrule
\multirow{3}{*}{Age Group} 
 & 61+        & 4,488 & 56.10 \\
 & 41–60      & 2,404 & 30.05 \\
 & 18–40      & 1,108 & 13.85  \\ 
\midrule
\multirow{2}{*}{Sex}
 & F          & 4,035 & 50.44 \\
 & M          & 3,965 & 49.56 \\
\bottomrule
\end{tabular}
\caption{Demographic Statistics of MIMIC-IV Data Upon Pre-processing.}
\label{tab:demographics_stat}
\end{table}

\subsection{Evaluation Protocols}
\label{sec:evaluation_protocols}

\paragraph{Inference setup.}
We evaluate two LLMs, \texttt{Gemini-2.0-Flash} \cite{team2023gemini} and \texttt{GPT-4o-mini} \cite{hurst2024gpt}. under a unified prompting framework (identical task description, input schema, and output format). Decoding is fixed with \texttt{temperature}=0.0 and \texttt{max\_tokens}=1024. Their context limits are: \texttt{Gemini-2.0-Flash} 1M tokens and \texttt{GPT-4o-mini} 128k tokens, thus the full prompt and visit notes fit without truncation.

\paragraph{Performance metrics.}
We report top-$k$ recall and precision. Especially, given an average of ~9 true diagnoses per patient, we set a single fixed $ k $ of 10. Next, to assess fairness, we compute True Positive Rate (TPR) and False Positive Rate (FPR), as these metrics capture disparities in correctly identifying conditions and in generating false alarms across groups. We also report the area under the precision–recall curve (AUPRC) for per-disease performance quantification.

\begin{figure*}[htbp]
    \centering
    \includegraphics[width=0.9\textwidth]{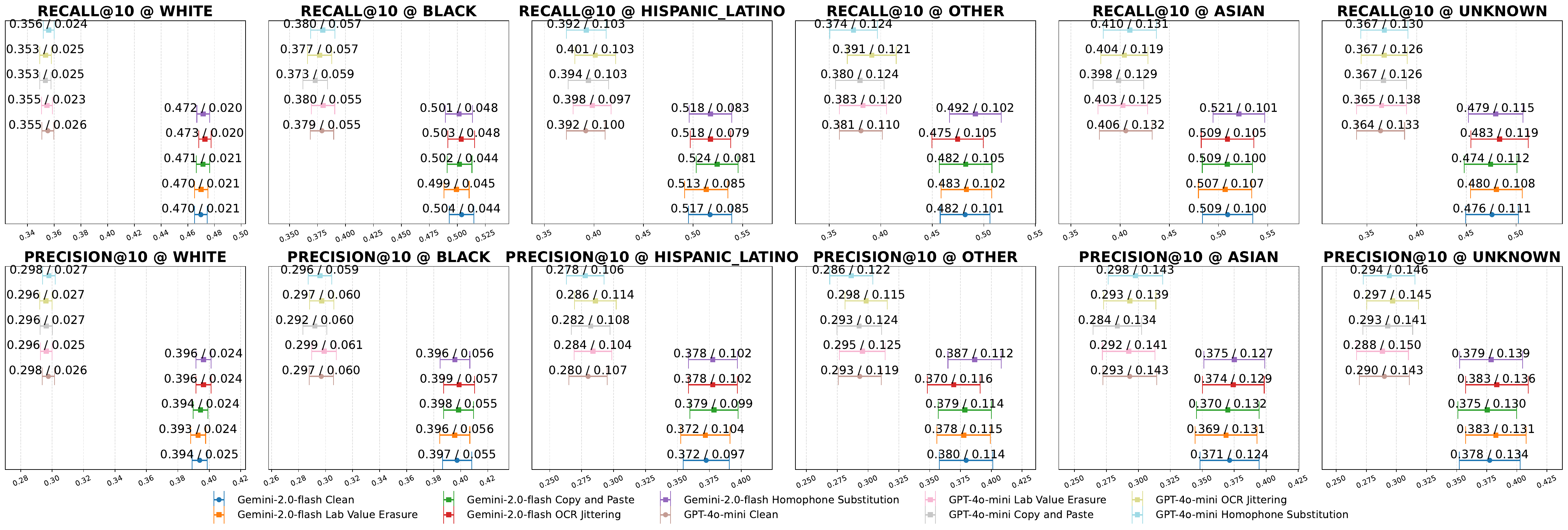}
    \caption{Performance across Racial Groups under Original and Corruption Settings.}
    \label{fig:performance_disparities_by_racial_subgroups}
\end{figure*}

Bootstrapping is used to evaluate the reliability of subgroup-specific estimates. By repeatedly resampling patient data, we generate empirical distributions of the metrics of interest. From these, 95\% confidence intervals are derived, whose relative widths indicate the degree of variability across subgroups. This approach quantifies the uncertainty in group-wise estimates and reveals potential disparities in robustness and fairness.

\section{Analysis and Discussions}
\label{sec:analysis_and_discussions}

\subsection{Overall Performance under Textual Degradation}
\label{sec:overall_performance}

\begin{table}[H]
\tiny
\centering
\setlength{\tabcolsep}{6pt}
\begin{tabular}{cccc}
\toprule
\midrule
\textbf{Model} & \textbf{Corruption}
& \textbf{Recall@10} 
& \textbf{Precision@10} \\
\hline
\midrule
\multirow{5}{*}{\texttt{Gemini-2.0-Flash}}
  & Original               & 0.4787 & 0.3918 \\
  & Lab-Value Erasure      & 0.4781 & 0.3908 \\
  & Prior-note Duplication & 0.4799 & 0.3924 \\
  & OCR Jittering          & 0.4808 & 0.3939 \\
  & Homophone Substitution & 0.4804 & 0.3940 \\
\midrule
\multirow{5}{*}{\texttt{GPT-4o-mini}}
  & Original               & 0.3629 & 0.2961 \\
  & Lab-Value Erasure      & 0.3630 & 0.2957 \\
  & Prior-note Duplication & 0.3608 & 0.2944 \\
  & OCR Jittering          & 0.3623 & 0.2957 \\
  & Homophone Substitution & 0.3636 & 0.2961 \\
\bottomrule
\end{tabular}
\caption{Performance under corruption settings. Mean values of Recall@10 and Precision@10 are reported.}
\label{tab:overall_table}
\end{table}

The results in Table \ref{tab:overall_table} indicate that \texttt{Gemini-2.0-Flash} consistently outperforms \texttt{GPT-4o-mini} across both Recall@10 and Precision@10, demonstrating its superior performance under varying corruption settings. Interestingly, the different forms of input corruption—such as lab-value omission, prior-note duplication, OCR jittering, and homophone substitution—tend to behave like mild data augmentation strategies, as they do not substantially degrade the overall retrieval performance. 

Nevertheless, while metrics convey stability, they do not provide sufficient insight into subgroup-level effects. In particular, the impact of such corruptions on minority groups remains uncertain, both in terms of performance and fairness. We therefore assess fairness to determine whether textual degradations disproportionately affect under-represented populations.

\subsection{Subgroup Diagnostic Disparities by Race}
\label{sec:subgroup_diagnostic_disparities_by_race}

\begin{figure}[htbp]
    \centering
    \includegraphics[width=0.7\columnwidth]{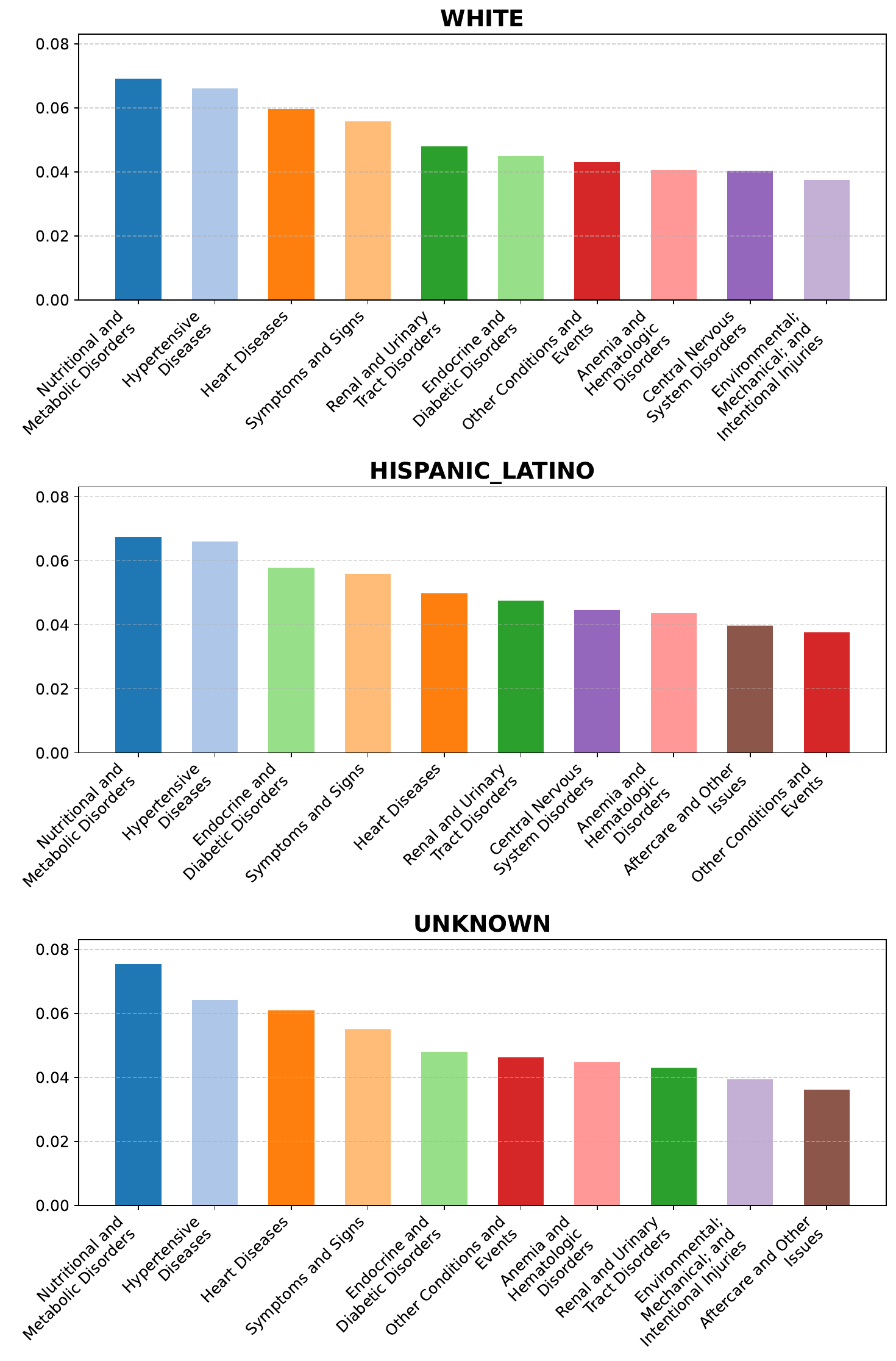}
    \caption{Top-10 Child-level Clinical Sub-categories across Race Groups (for simplicity, we report results for three groups: White, Hispanic/Latino, and Unknown).}
    \label{fig:diagnostic_disparities_by_race_subgroups}
\end{figure}

\begin{figure*}[htbp]
    \centering
    \includegraphics[width=0.9\textwidth]{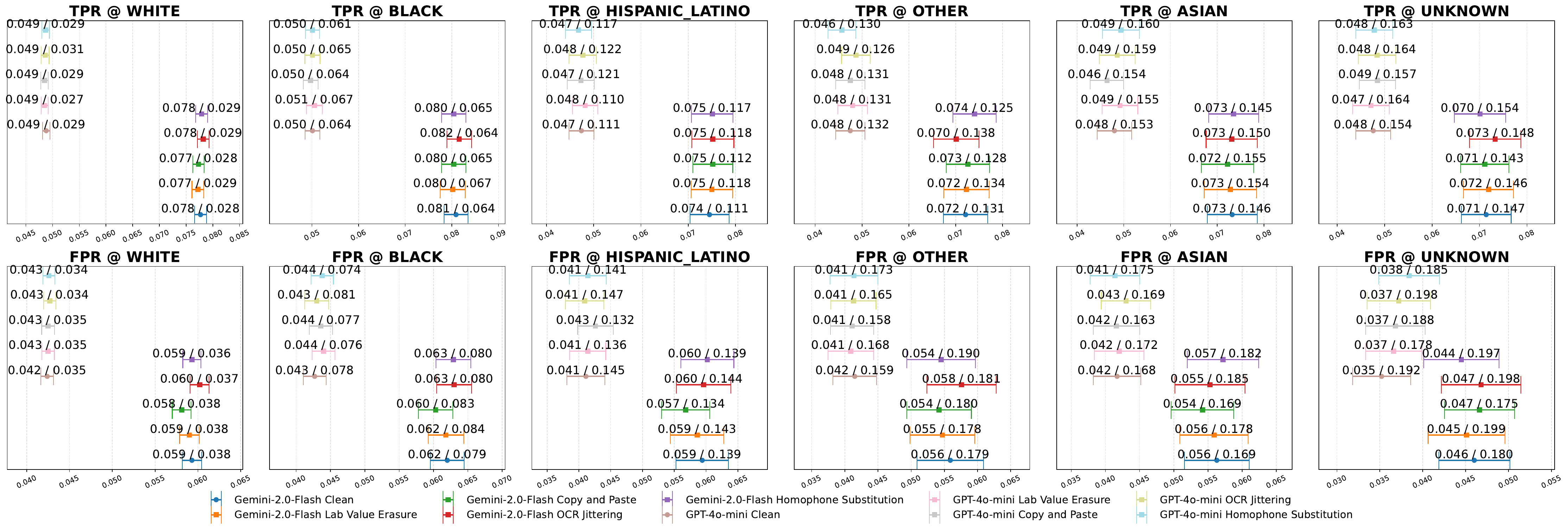}
    \caption{Fairness across Racial Groups under Original and Corruption Settings.}
    \label{fig:fairness_disparities_by_racial_subgroups}
\end{figure*}

\paragraph{Clinical Heterogeneity Across Racial Groups.} As provided in Figure \ref{fig:diagnostic_disparities_by_race_subgroups}, white patients most often presented with nutritional, metabolic, hypertensive, and heart diseases, as well as renal, endocrine, and diabetic disorders. Hispanic/Latino patients showed a similar cardiometabolic profile but with heightened representation of mood, anxiety, and cognitive disorders. Patients with Unknown race labels demonstrated comparatively greater frequencies of injury-related categories and other external causes, coupled with substantial renal and metabolic conditions, reflecting demographic and documentation disparities in clinical coding.

\paragraph{Performance Analysis.} As shown in Figure \ref{fig:performance_disparities_by_racial_subgroups}, minority racial subgroups exhibit heightened volatility in predictive metrics when subjected to textual corruptions. Their recall@10 and precision@10 values not only fluctuate more substantially than those observed in the White cohort but also present markedly broader bootstrap uncertainty intervals. These discrepancies persist even after controlling for cohort size, implying that instability is not merely a function of smaller representation. Instead, they underscore that current models generalise less reliably to minority populations, rendering these groups disproportionately susceptible to degradation under textual perturbations and amplifying concerns about clinical equity in deployment.

\paragraph{Fairness Analysis.} Fairness metrics corroborate the instability observed in predictive performance (Figure \ref{fig:fairness_disparities_by_racial_subgroups}). True Positive Rate (TPR) and False Positive Rate (FPR) vary substantially more for minority groups under corruption, with both means and error bars showing greater volatility. This instability cannot be explained away by data scarcity, but instead points to systematic fragility in how errors are distributed across subgroups. Such sensitivity implies that models which appear equitable under clean conditions may mask significant disparities under realistic perturbations, underscoring the need to evaluate subgroup stability as a central aspect of fairness. A detailed version of the table is included in the Appendix \ref{sec:appendix_disparities_by_racial_subgroups}.

\subsection{Subgroup Diagnostic Disparities by Age Groups}
\label{sec:subgroup_diagnostic_disparities_by_age}

\begin{figure}[htbp]
    \centering
    \includegraphics[width=0.7\columnwidth]{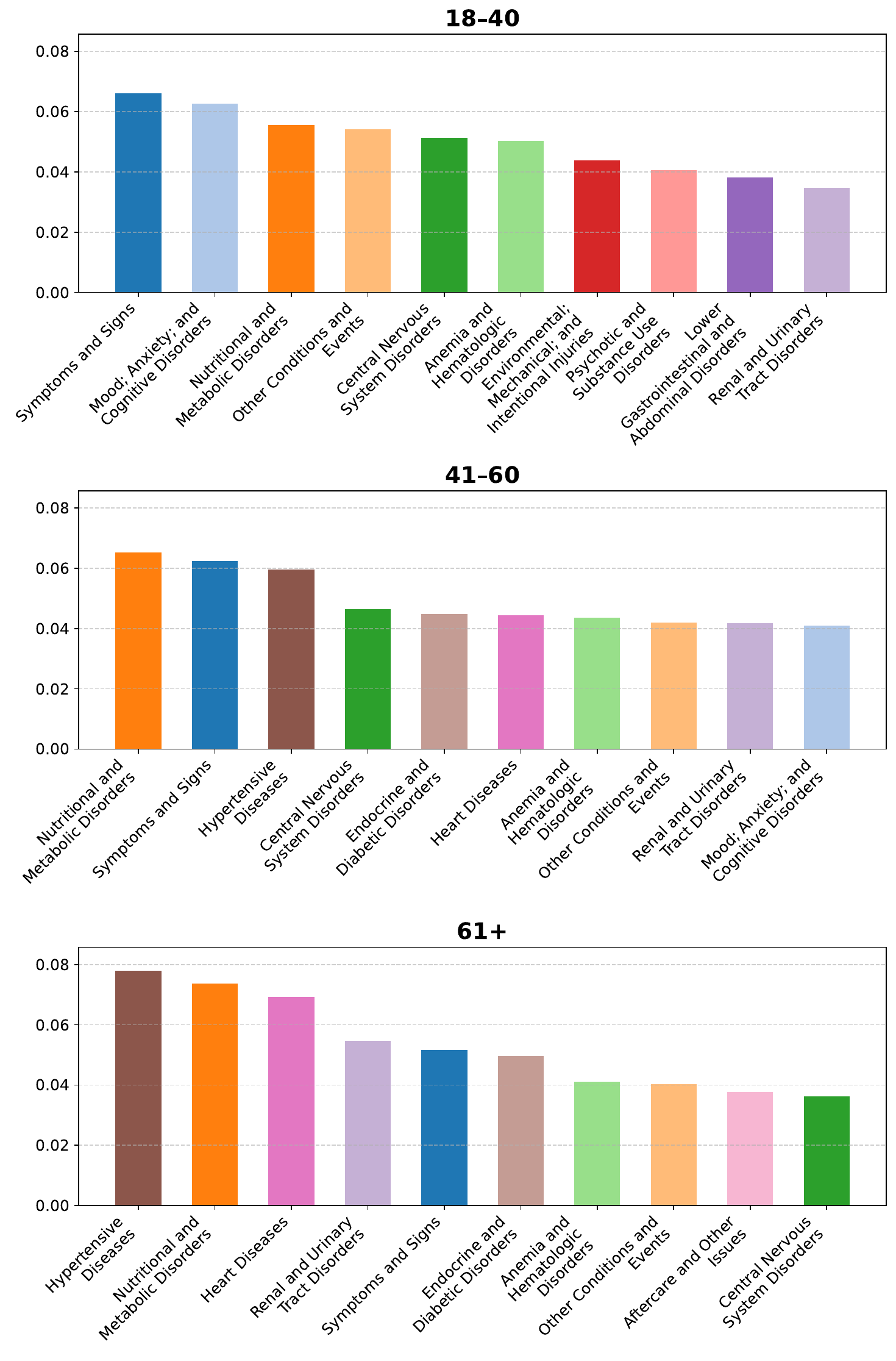}
    \caption{Top-10 Child-level Clinical Sub-categories across Age Groups (18–40, 41–60, 61+).}
    \label{fig:diagnostic_disparities_by_age_subgroups}
\end{figure}

\paragraph{Clinical Heterogeneity Across Age Groups.} Younger adults (18–40 years) most frequently presented with mood, anxiety, and cognitive disorders, symptoms-based visits, and injury-related diagnoses. Middle-aged groups (41–60 years) displayed a shift toward metabolic, hypertensive, and endocrine or diabetic diseases, with emerging renal and cardiovascular comorbidities. In older adults (61+ years), hypertensive, heart, renal, and metabolic disorders dominated the diagnostic spectrum, emphasising age-dependent progression from psychosomatic to chronic systemic conditions. We attach the regarding Figure \ref{fig:diagnostic_disparities_by_age_subgroups}.

\begin{figure*}[htbp]
    \centering
    \includegraphics[width=0.9\textwidth]{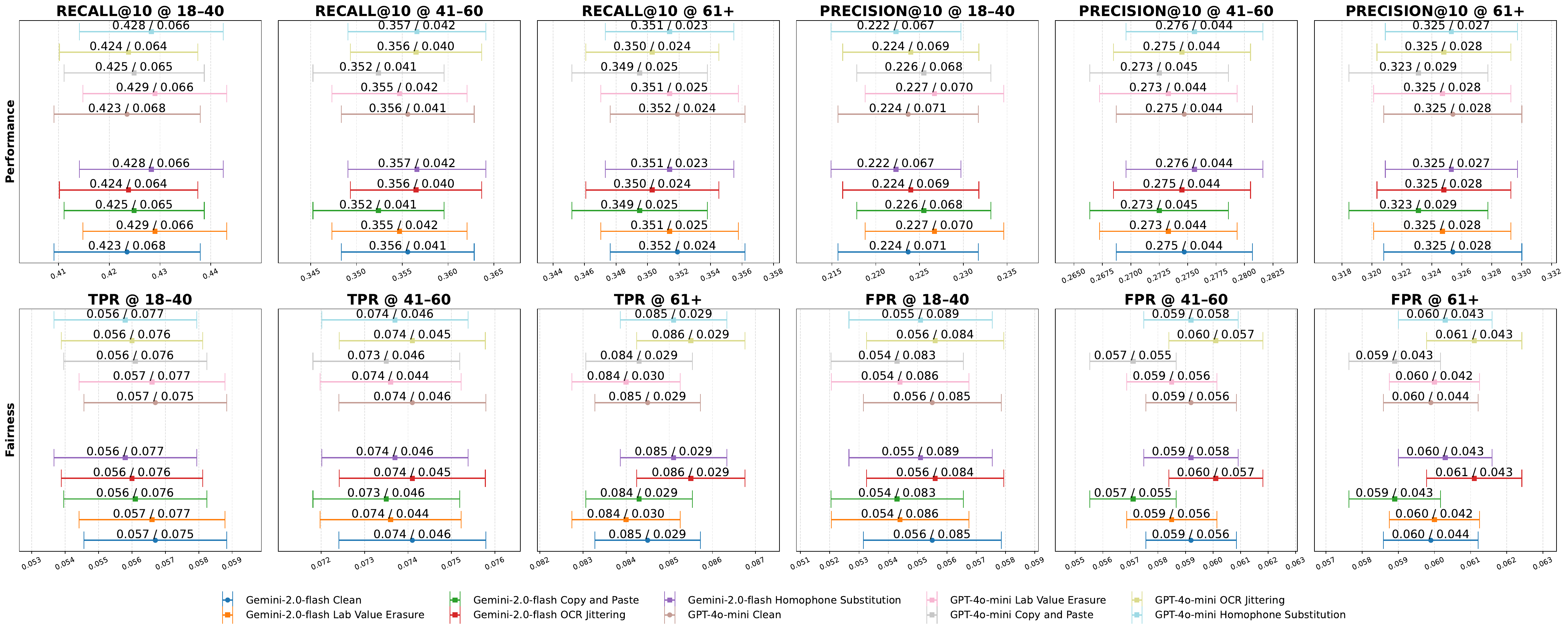}
    \caption{Performance and Fairness across Age Groups under Original and Corruption Settings.}
    \label{fig:disparities_by_age_subgroups}
\end{figure*}

\paragraph{Performance Analysis.}
Performance across age groups revealed an expected pattern of robustness. Older patients (61+) exhibited greater stability under textual corruptions, maintaining consistent performance with narrower error bounds. Their performance remained steady even under severe perturbations. In contrast, younger patients (18–40) showed notable volatility, with irregular fluctuations and wider uncertainty intervals. These trends suggest that instability in younger groups likely stems from training data imbalance rather than sample size limitations.

\paragraph{Fairness Analysis.}
Fairness metrics echoed this age-dependent resilience. Among older patients (61+), TPR and FPR remained balanced and stable across corruptions, while younger cohorts (18–40) exhibited greater variability in both metrics. The middle-aged group again fell between these extremes. Thus, fairness for older populations appears more robust to corruptions, whereas younger subgroups face disproportionate degradation, emphasising the need to boost subgroup stability under realistic text perturbations. The complete table is provided in the Appendix \ref{sec:appendix_disparities_by_age_subgroups}.

\subsection{Sub-group Diagnostic Disparities by Sex on Diagnosis Prediction}
\label{sec:appendix_diagnosis_sex}

\paragraph{Clinical Heterogeneity Across Sex Groups.} 

\begin{figure}[htbp]
    \centering
    \includegraphics[width=0.7\columnwidth]{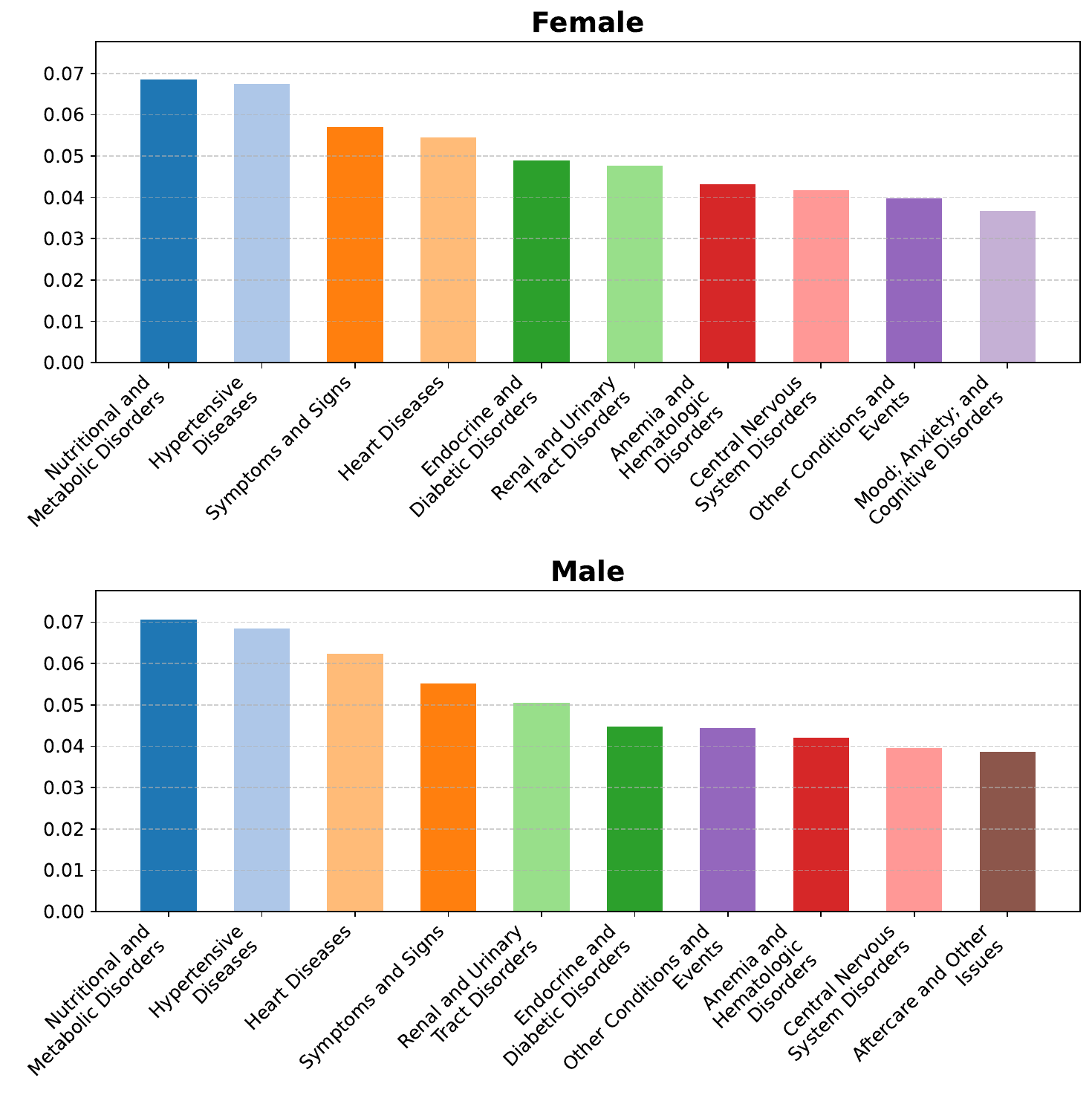}
    \caption{Top-10 Child-level Clinical Sub-categories across Sex Groups (Female and Male).}
    \label{fig:diagnostic_disparities_by_sex_subgroups}
\end{figure}

\begin{figure}[htbp]
    \centering
    \includegraphics[width=1.0\columnwidth]{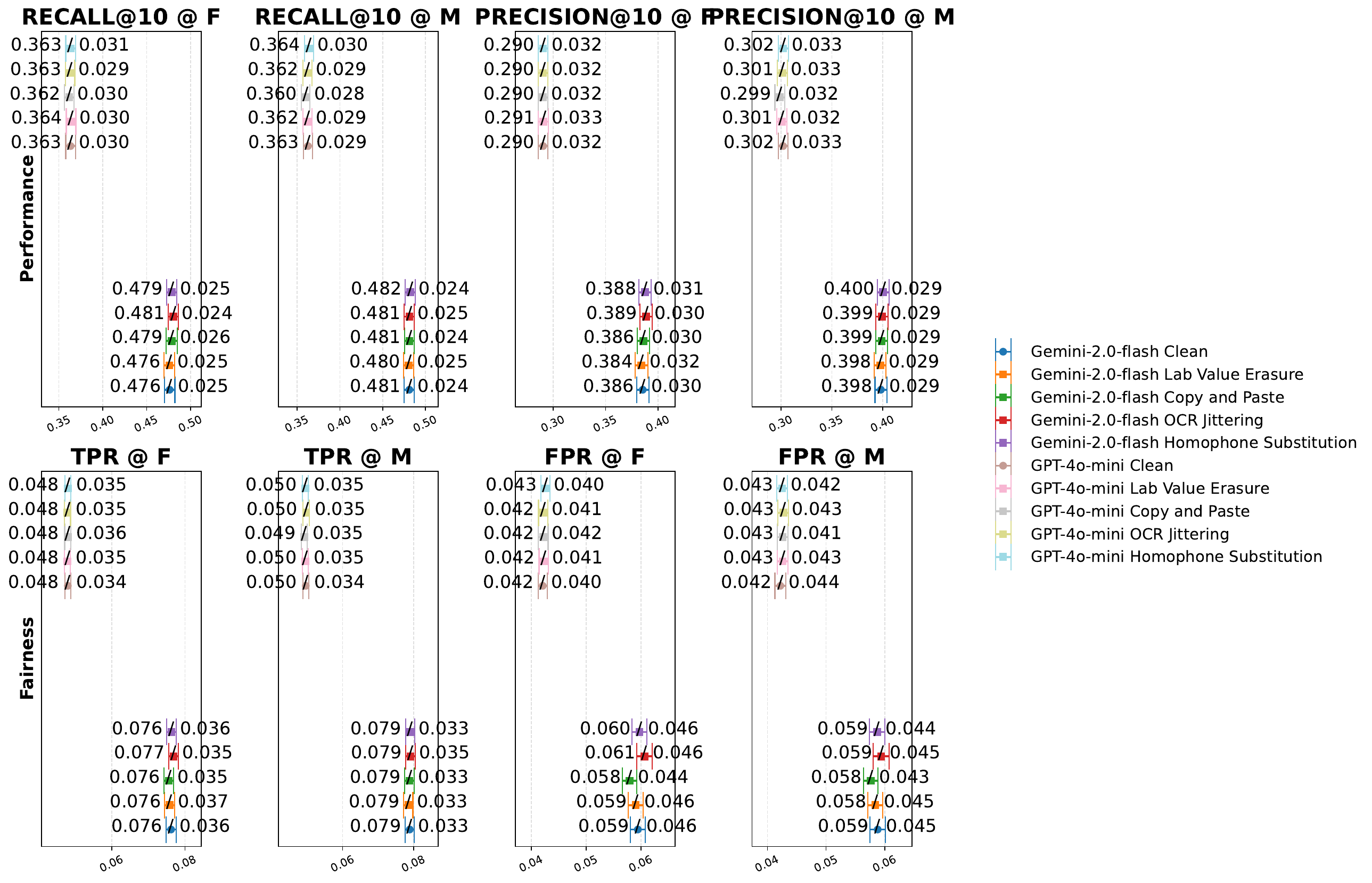}
    \caption{Performance and Fairness across Sex Groups under Original and Corruption Settings.}
    \label{fig:disparities_by_sex_subgroups}
    \label{fig:sex}
\end{figure}

Both male and female patients most commonly exhibited nutritional and metabolic disorders, hypertensive and heart diseases, renal and urinary tract disorders, and endocrine or diabetic conditions. Females showed a relatively higher proportion of mood, anxiety, and cognitive disorders, alongside aftercare and symptomatic visits. In contrast, males had slightly greater representation in cardiovascular and renal conditions, reflecting sex-linked variations in cardiometabolic and neuropsychiatric disease burden.

\paragraph{Performance and Fairness Analysis.} 

Across sex subgroups, both performance and fairness metrics displayed remarkable consistency under textual corruption. Female and male patients achieved nearly identical recall@10 and precision@10 values with minimal fluctuations and tight confidence intervals across all perturbation types. Similarly, TPR and FPR remained stable and closely aligned between groups, indicating balanced error distributions and robust generalisation. These results suggest that the model’s behavior is relatively equitable and resilient across sexes, with no evident subgroup disproportionately affected by corruption-induced degradation—contrasting with the disparities observed along other demographic dimensions.

\begin{table*}[t]
\centering
\scriptsize
\begin{tabular}{ccccccccc}
\toprule
\multirow{2}{*}{\textbf{Group}} 
& \multirow{2}{*}{\textbf{Subgroup}} 
& \multirow{2}{*}{\textbf{Criteria}} 
& \multirow{2}{*}{\textbf{Settings}} 
& \multirow{2}{*}{\textbf{NECHO v3 (Full)}} 
& \multicolumn{3}{c}{\textbf{Ablation Studies}} 
& \textbf{Comparative Study} \\
\cmidrule(lr){6-8}\cmidrule(lr){9-9}
& & & & & \textbf{w/o (i)} & \textbf{w/o (ii)} & \textbf{w/o (iii)} 
& \textbf{Parental-level} \\
\midrule
\multirow{2}{*}{Overall} 
   & \multirow{2}{*}{--} & \multirow{2}{*}{Performance} 
    & Original & \textbf{0.3629} & 0.3580 & 0.3425 & 0.1085 & 0.9874 \\
  & & & Corruption & \textbf{0.3623} & 0.3571 & 0.3473 & 0.0957 & 0.9870 \\
\midrule
\multirow{8}{*}{Subgroup} 
  & \multirow{4}{*}{Unknown} 
    & \multirow{2}{*}{Performance} 
      & Original & \textbf{0.3637} & 0.3597 & 0.3534 & 0.1016 & 0.9865 \\
  & & & Corruption & \textbf{0.3673} & 0.3606 & 0.3529 & 0.0942 & 0.9861 \\
  &  & \multirow{2}{*}{Fairness} 
      & Original & \textbf{0.0488} & 0.0486 & 0.0474 & 0.0188 & 0.0525 \\
  & & & Corruption & \textbf{0.0486} & 0.0471 & 0.0465 & 0.0175 & 0.0651 \\
\cmidrule(lr){2-9}
  & \multirow{4}{*}{18--40} 
    & \multirow{2}{*}{Performance} 
      & Original & \textbf{0.4235} & 0.4173 & 0.4180 & 0.1224 & 0.9868 \\
  & & & Corruption & \textbf{0.4238} & 0.4188 & 0.4112 & 0.1135 & 0.9864 \\
  &  & \multirow{2}{*}{Fairness} 
      & Original & \textbf{0.0365} & 0.0354 & 0.0343 & 0.0163 & 0.0528 \\
  & & & Corruption & \textbf{0.0365} & 0.0357 & 0.0351 & 0.0151 & 0.0650 \\
\bottomrule
\end{tabular}
\caption{Ablation and comparative studies on overall and subgroup-specific (Unknown, 18--40) performance and fairness. The three ablations correspond to (i) the hierarchical chain-of-thought (CoT), (ii) demographic information, and (iii) the proposed label mapping strategy.}
\label{tab:ablation_studies}
\end{table*}
\vspace{-10pt}

\subsection{Ablation and Comparative Studies}
\label{sec:ablation_studies}

Table \ref{tab:ablation_studies} summarises the ablation results for three key components of our framework: hierarchical CoT, demographic information, and label mapping. Using \texttt{GPT-4o-mini} under OCR jittering—the most impactful corruption in our evaluations—we observe that the full model consistently outperforms all ablated variants in both accuracy and subgroup fairness. Demographic information improves calibration under noise, whereas removing label mapping markedly degrades performance. The strong parental-level results further support the value of hierarchical CoT. Overall, the three components jointly enable robust and equitable modelling.

\subsection{Long-tailedness for Minority Subgroups}
\label{sec:longtailedness}

\begin{table}[t]
\centering
\scriptsize
\setlength{\tabcolsep}{3pt}
\begin{tabular}{cccccc}
\toprule
\midrule
\textbf{Disease} & \textbf{Original} & \textbf{Omission} & \textbf{Duplication} & \textbf{OCR} & \textbf{Homophone} \\
\hline
\midrule
\multicolumn{6}{c}{\textbf{Racial Group: Unknown}} \\
\midrule
\makecell[l]{Cerebrovascular \\ Disorders} & 0.1521 & 0.1553 & 0.1635 & 0.1517 & 0.1457 \\
\makecell[l]{Hematologic and \\ Endocrine Cancers} & 0.1927 & 0.1843 & 0.1903 & 0.1866 & 0.1918 \\
\makecell[l]{Gastrointestinal \\ Cancers} & 0.2024 & 0.2218 & 0.1977 & 0.2173 & 0.2057 \\
\midrule
\multicolumn{6}{c}{\textbf{Age Group: 18--40}} \\
\midrule
\makecell[l]{Hematologic and \\ Endocrine Cancers} & 0.3646 & 0.2092 & 0.3074 & 0.2030 & 0.2304 \\
\makecell[l]{Cancers of \\ Other Systems} & 0.1342 & 0.1001 & 0.0903 & 0.1772 & 0.1137 \\
\makecell[l]{Postpartum and \\ Puerperal \\ Complications} & 0.0335 & 0.0335 & 0.0335 & 0.0334 & 0.0334 \\
\bottomrule
\end{tabular}
\caption{Per-disease AUPRC under different corruption settings for minority subgroups based on race (Unknown) and age (18--40) using \texttt{GPT-4o-mini}.}
\label{tab:per_disease_auprc}
\end{table}

Table \ref{tab:per_disease_auprc} presents the per-disease AUPRC for minority subgroups under different corruption scenarios using \texttt{GPT-4o-mini}. Long-tailed diagnoses (tail classes) such as Physical Trauma and Injuries and Reproductive Disorders among the 18–40 age group (AUPRC $<$ 0.20), as well as Cerebrovascular Disorders and Gastrointestinal Cancers among patients with unknown race (AUPRC $<$ 0.40), exhibit persistently low predictive performance. These findings suggest that infrequent clinical conditions remain challenging to model even under clean inputs, and that textual corruption further amplifies the performance disparities observed in under-represented subpopulations.

\section{Conclusion}
\label{sec:conclusion}

We evaluate LLMs for next-visit diagnosis prediction using the MIMIC-IV dataset, focusing on resilience to textual corruption. Despite stable overall performance, significant variability emerges in minority subgroups and tail diagnoses, posing fairness risks for clinical deployment. To mitigate these issues, we introduce NECHO v3, which integrates label reduction with hierarchical CoT reasoning to enhance robustness and equity under degraded texts.

Our study has several limitations. First, the analysis relies on a single dataset, which may not generalise to other clinical settings. Second, we use zero-shot evaluation, whereas real-world systems often incorporate local fine-tuning that could shift performance and fairness outcomes. Third, real-world textual noise typically arises from interacting sources, while our corruptions are assessed independently. We encourage future work to address these constraints.

\section{Ethical Considerations}
\label{sec:ethical_consirations}

This study highlights that textual degradations increase LLM uncertainty and may unevenly affect demographic subgroups, potentially reinforcing existing inequities in clinical decision support. While our approach mitigates some of these risks through improved robustness and fairness, responsible deployment requires ongoing monitoring and adherence to institutional governance. LLM-assisted predictions should complement—rather than replace—clinical judgement, especially where heightened uncertainty could disproportionately impact vulnerable populations.

\newpage

\bibliography{aaai2026}

\newpage

\onecolumn
\section{Prompt}
\label{sec:appendix_prompt}

\begin{tcolorbox}[nobeforeafter, 
    title=An example of a complete prompt template for next visit diagnosis prediction, 
    colframe=darkgray,
    colback=white, 
    breakable]
\small
    
You are an experienced critical care physician working in an Intensive Care Unit (ICU). You are skilled in interpreting ICU discharge summaries of patients, and predicting clinical outcomes based on the patient's status at the time of discharge and their overall hospital course.\\

Your primary task is to assess the patient’s complete longitudinal ICU record—**all prior and current ICU discharge summaries together with basic demographic information**—to predict which diagnosis categories (from the predefined candidate set) are most likely to be coded at the patient’s next hospital visit.\\

The diagnostic taxonomy is hierarchical. Use the following structure:\\

**Parent diagnostic systems** (e.g., "Diseases of the Circulatory System")  \\
→ **Child-level clinical subcategories** (e.g., "Cardiac Conditions", "Cerebrovascular Disorders") \\

Use the following clinical features to guide your prediction:\\

- Longitudinal comorbidities (e.g., CHF, diabetes, COPD)  \\
- Acute-on-chronic exacerbations and organ failure  \\
- New or recurrent infections  \\
- ICU interventions (e.g., ventilation, dialysis, vasopressors)  \\
- Functional decline or unresolved complications at discharge  \\
- Trends in labs, imaging, or organ support  \\
- Discharge disposition (e.g., home, rehab, hospice)\\

---

\#\#\# Candidate Parent Diagnostic Systems

- Perinatal and Congenital Conditions \\
- Diseases of the Blood and Immune System \\
- Diseases of the Circulatory System   \\
- Diseases of the Respiratory System   \\
- Diseases of the Digestive System   \\
- Diseases of the Genitourinary System   \\
- Pregnancy; Childbirth; and Postpartum Complications   \\
- Diseases of the Musculoskeletal and Connective Tissue   \\
- Diseases of the Nervous System and Sense Organs   \\
- Endocrine; Nutritional; and Metabolic Diseases  \\
- Infectious and Parasitic Diseases   \\
- Diseases of the Skin and Subcutaneous Tissue   \\
- Injury; Poisoning; and External Causes   \\
- Symptoms, Signs, and Other Conditions \\
- Mental; Behavioral; and Neurodevelopmental Disorder   \\
- Neoplasms   \\
- External Causes of Morbidity (E-Codes) \\

--- \\
\#\#\# Allowed Child Subcategories \\

You must select one or more child-level subcategories **only** from this predefined mapping: \\

\{ \\
  "Perinatal and Congenital Conditions": [ \\
    "Neonatal Trauma and Injury", \\
    "Other Perinatal Conditions", \\
    "Congenital Anomalies" \\
  ], \\
  "Diseases of the Blood and Immune System": [ \\
    "Anemia and Hematologic Disorders", \\
    "Immunologic Disorders" \\
  ], \\
  ... (Omitted for brevity) \\
\} \\

For each selected **parent-level diagnostic system**, you **MUST** select **one or more** clinically plausible child-level diagnosis categories. If the patient’s record supports multiple complications, sequelae, or chronic comorbidities within a system, include **all applicable child categories**. Avoid under-selecting. Draw upon clinical history, trends, and discharge markers. \\

---

\#\#\# Step-by-Step Reasoning Process \\

1. **Timeline Synthesis**: Describe how the patient’s health has evolved across ICU encounters (e.g., persistent CHF, recent sepsis, renal decline).   \\
2. **Diagnostic Trait Inference**: Infer clinical risks based on underlying disease traits or complications (e.g., steroid use → infection risk, multiple admissions → critical illness).   \\
3. **Parent-Level Filtering**: Select all high-level diagnostic systems relevant to the patient’s longitudinal profile.   \\
4. **Subcategory Disambiguation**: For each selected parent, choose **one or more** matching child-level categories, based strictly on the allowed schema.   \\
5. **Prediction**: Output a JSON dictionary mapping each parent to its most plausible child categories. \\

---

\#\#\# Output Format \\

Return a single JSON object with **exactly two keys**: \\

1. `"think"`: A clinical reasoning narrative (less than 150 words), including timeline synthesis, pathophysiological reasoning, and justification for each selected parent and subcategory.   \\
2. `"answer"`: A dictionary mapping each predicted **parent-level diagnostic system** to a **list of one or more** predicted child-level categories. \\

---

\#\#\# Example Output: \\
```json\\
\{
  "think": "This patient has had multiple ICU admissions over 18 months for acute decompensated heart failure with reduced ejection fraction and atrial fibrillation, complicated by pulmonary edema and cardiogenic shock requiring inotropes. Renal function shows progressive chronic kidney disease with episodes of acute tubular injury during volume shifts. The course has been further complicated by recurrent MRSA pneumonia necessitating intubation and prolonged antibiotics, as well as chronic steroid use for autoimmune vasculitis. At the most recent discharge, the patient had unresolved fluid overload and borderline oxygenation and was transferred to a skilled nursing facility. Overall, the patient remains at high risk for recurrent cardiac decompensation, infectious complications, worsening renal failure, and continued critical illness.",\\
  "answer": {\\
    "Diseases of the Circulatory System": [
      "Heart Diseases",
      "Peripheral and Venous Diseases"
    ],\\
    "Diseases of the Respiratory System": [
        "Respiratory Infections"
    ],\\
    "Diseases of the Genitourinary System": [
      "Renal and Urinary Tract Disorders"
    ],\\
    "Infectious and Parasitic Diseases": [
      "Bacterial and Septic Infections",
      "Viral; Mycotic; and Other Infections"
    ]
  }
\}
'''

\end{tcolorbox}

\section{Diagnosis Mapping}
\label{sec:appendix_diagnosis_mapping}

\small
\begin{longtable}{p{5cm}|p{5cm}|p{6cm}}

\hline
\midrule
\textbf{Parent-Level Diagnostic Systems} & \textbf{Child-level Clinical Sub-categories} & \textbf{Specific Diagnoses} \\
\hline
\midrule
\endfirsthead
\hline
\midrule
\textbf{Parent-Level Diagnostic Systems} & \textbf{Child-level Clinical Sub-categories} & \textbf{Specific Diagnoses} \\
\hline
\midrule
\endhead

\multirow{3}{5cm}{Perinatal and Congenital Conditions} &
Neonatal Trauma and Injury &
\begin{tabular}[t]{@{}l@{}}
Fracture of skull or face\\
Fracture of arm\\
Fracture of leg\\
Other fracture\\
Birth trauma\\
Joint injury\\
Spinal cord injury
\end{tabular} \\
\cline{2-3}
& Other Perinatal Conditions &
\begin{tabular}[t]{@{}l@{}}
Low birth weight\\
Perinatal jaundice\\
Birth asphyxia\\
Liveborn infant\\
Other perinatal diseases
\end{tabular} \\
\cline{2-3}
& Congenital Anomalies &
\begin{tabular}[t]{@{}l@{}}
Nervous system congenital anomaly\\
Fracture of hip\\
Other congenital anomalies\\
Gastrointestinal congenital anomaly\\
Genitourinary congenital anomaly\\
Cardiac anomalies
\end{tabular} \\
\hline

\multirow{2}{5cm}{Diseases of the Blood and Immune System} &
Anemia and Hematologic Disorders &
\begin{tabular}[t]{@{}l@{}}
Anemia\\
Acute post-hemorrhagic anemia\\
Sickle cell\\
Coagulation and hemorrhagic disorders\\
White blood cell disorders\\
Other hematologic diseases
\end{tabular} \\
\cline{2-3}
& Immunologic Disorders &
\begin{tabular}[t]{@{}l@{}}
Immunity disorders
\end{tabular} \\
\hline

\multirow{4}{5cm}{Diseases of the Circulatory System} &
Heart Diseases &
\begin{tabular}[t]{@{}l@{}}
Heart valve disease\\
Carditis\\
Acute myocardial infarction\\
Coronary atherosclerosis\\
Pulmonary heart disease\\
Other heart diseases\\
Conduction disorders\\
Chest pain\\
Cardiac dysrhythmia\\
Cardiac arrest\\
Congestive heart failure; non-hypertensive
\end{tabular} \\
\cline{2-3}
& Hypertensive Diseases &
\begin{tabular}[t]{@{}l@{}}
Hypertension\\
Hypertension complications
\end{tabular} \\
\cline{2-3}
& Cerebrovascular Disorders &
\begin{tabular}[t]{@{}l@{}}
Acute cerebrovascular disease\\
Pre-cerebral occlusion\\
Other cerebrovascular diseases\\
Transient ischemic attack\\
Late effects of cerebrovascular disease
\end{tabular} \\
\cline{2-3}
& Peripheral and Venous Diseases &
\begin{tabular}[t]{@{}l@{}}
Peripheral atherosclerosis\\
Aneurysm\\
Arterial embolism\\
Other circulatory diseases\\
Phlebitis\\
Varicose vein\\
Other vein disorders
\end{tabular} \\
\hline

\multirow{3}{5cm}{Diseases of the Respiratory System} &
Respiratory Infections &
\begin{tabular}[t]{@{}l@{}}
Pneumonia\\
Influenza\\
Tonsillitis\\
Bronchitis\\
Other upper respiratory infections
\end{tabular} \\
\cline{2-3}
& Chronic and Obstructive Pulmonary Diseases &
\begin{tabular}[t]{@{}l@{}}
Chronic obstructive pulmonary disease\\
Asthma
\end{tabular} \\
\cline{2-3}
& Other Respiratory Conditions &
\begin{tabular}[t]{@{}l@{}}
Aspiration pneumonia\\
Pleurisy\\
Respiratory distress\\
Adult respiratory failure\\
Lung diseases due to external agents\\
Other lower respiratory diseases\\
Other upper respiratory diseases
\end{tabular} \\
\hline

\multirow{4}{5cm}{Diseases of the Digestive System} &
Upper Gastrointestinal Disorders &
\begin{tabular}[t]{@{}l@{}}
Esophageal disease\\
Gastric and duodenal ulcer\\
Gastritis\\
Other diseases of stomach\\
Gastroenteritis\\
Gastrointestinal hemorrhage
\end{tabular} \\
\cline{2-3}
& Lower Gastrointestinal and Abdominal Disorders &
\begin{tabular}[t]{@{}l@{}}
Appendicitis\\
Ulcerative colitis\\
Intestinal obstruction\\
Diverticulosis\\
Anal and rectal conditions\\
Abdominal hernia\\
Peritonitis\\
Hemorrhoids\\
Other gastrointestinal diseases
\end{tabular} \\
\cline{2-3}
& Hepatic and Pancreatic Disorders &
\begin{tabular}[t]{@{}l@{}}
Other liver diseases\\
Pancreas disease\\
Biliary disease
\end{tabular} \\
\cline{2-3}
& Oral and Dental Conditions &
\begin{tabular}[t]{@{}l@{}}
Teeth diseases\\
Mouth diseases
\end{tabular} \\
\hline

\multirow{2}{5cm}{Diseases of the Genitourinary System} &
Renal and Urinary Tract Disorders &
\begin{tabular}[t]{@{}l@{}}
Nephritis\\
Acute renal failure\\
Chronic kidney disease\\
Urinary tract infection\\
Urinary stone\\
Other diseases of kidney\\
Other diseases of bladder\\
Other genitourinary diseases
\end{tabular} \\
\cline{2-3}
& Reproductive Disorders &
\begin{tabular}[t]{@{}l@{}}
Pelvic inflammatory disease\\
Endometriosis\\
Prolapse\\
Menstrual disorders\\
Ovarian cyst\\
Menopausal disorders\\
Female infertility\\
Breast disease\\
Contraceptive management\\
Other female genital disorders\\
Infections of male genital organs\\
Other male genital disorders\\
Benign prostatic hyperplasia
\end{tabular} \\
\hline

\multirow{3}{5cm}{Pregnancy; Childbirth; and Postpartum Complications} &
Pregnancy Complications &
\begin{tabular}[t]{@{}l@{}}
Induced abortion\\
Spontaneous abortion\\
Abortion complications\\
Ectopic pregnancy\\
Long pregnancy\\
Early labor\\
Malposition of fetus\\
Pelvic obstruction\\
Diabetes mellitus in pregnancy\\
Hypertension in pregnancy\\
Hemorrhage in pregnancy\\
Previous C-section\\
Other pregnancy complications
\end{tabular} \\
\cline{2-3}
& Labor and Delivery Complications &
\begin{tabular}[t]{@{}l@{}}
Fetal distress\\
Amniotic fluid disorders\\
Umbilical cord complications\\
Obstetrics-related perinatal trauma\\
Forceps delivery\\
Other complications of birth\\
Other pregnancy and delivery \\ \hspace*{2em} including normal
\end{tabular} \\
\cline{2-3}
& Postpartum and Puerperal Complications &
\begin{tabular}[t]{@{}l@{}}
Osteoarthros\\
Other joint diseases
\end{tabular} \\
\hline

\multirow{2}{5cm}{Diseases of the Musculoskeletal and Connective Tissue} &
Autoimmune and Connective Tissue Disorders &
\begin{tabular}[t]{@{}l@{}}
Systemic lupus erythematosus\\
Infectious arthritis\\
Rheumatoid arthritis\\
Other connective tissue disorders
\end{tabular} \\
\cline{2-3}
& Skeletal and Acquired Musculoskeletal Disorders &
\begin{tabular}[t]{@{}l@{}}
Osteoporosis\\
Other bone diseases\\
Pathological fracture\\
Back problem\\
Acquired foot deformity
\end{tabular} \\
\hline

\multirow{2}{5cm}{Diseases of the Nervous System and Sense Organs} &
Central Nervous System Disorders &
\begin{tabular}[t]{@{}l@{}}
Epilepsy and convulsive disorders\\
Meningitis\\
Encephalitis\\
Other CNS infections\\
Parkinson's disease\\
Multiple sclerosis\\
Other hereditary central nervous system \\ \hspace*{1em} diseases\\
Other nervous system diseases\\
Headache and migraine\\
Coma and brain damage\\
Paralysis
\end{tabular} \\
\cline{2-3}
& Sensory and Vestibular Disorders &
\begin{tabular}[t]{@{}l@{}}
Eye infection\\
Cataract\\
Retinal disease\\
Glaucoma\\
Other eye diseases\\
Otitis media\\
Dizziness\\
Other ear diseases
\end{tabular} \\
\hline

\multirow{2}{5cm}{Endocrine; Nutritional; and Metabolic Diseases} &
Endocrine and Diabetic Disorders &
\begin{tabular}[t]{@{}l@{}}
Thyroid disorders\\
Diabetes Mellitus without complications\\
Diabetes Mellitus with complications\\
Other endocrine disorders
\end{tabular} \\
\cline{2-3}
& Nutritional and Metabolic Disorders &
\begin{tabular}[t]{@{}l@{}}
Nutritional deficiencies\\
Hyperlipidemia\\
Gout and other crystal arthropathies\\
Fluid and electrolyte disorders\\
Cystic fibrosis\\
Other nutritional and metabolic disorders
\end{tabular} \\
\hline

\multirow{3}{5cm}{Infectious and Parasitic Diseases} &
Bacterial and Septic Infections &
\begin{tabular}[t]{@{}l@{}}
Tuberculosis\\
Septicemia\\
Intestinal infection\\
Other bacterial infections
\end{tabular} \\
\cline{2-3}
& Viral; Mycotic; and Other Infections &
\begin{tabular}[t]{@{}l@{}}
HIV infection\\
Hepatitis\\
Viral infection\\
Mycoses\\
Other infectious diseases
\end{tabular} \\
\cline{2-3}
& Sexually Transmitted and Preventive Conditions &
\begin{tabular}[t]{@{}l@{}}
Sexual Infections\\
Immunization and screening
\end{tabular} \\
\hline

\multirow{1}{5cm}{Diseases of the Skin and Subcutaneous Tissue} &
Inflammatory and Infectious Skin Disorders &
\begin{tabular}[t]{@{}l@{}}
Skin infection\\
Skin ulcer\\
Other skin diseases\\
Other inflammatory skin conditions\\
Other acquired deformities
\end{tabular} \\
\hline

\multirow{2}{5cm}{Injury; Poisoning; and External Causes} &
Physical Trauma and Injuries &
\begin{tabular}[t]{@{}l@{}}
Sprain\\
Intracranial injury\\
Crush injury\\
Open wound of head\\
Open wound of extremity\\
Superficial injury\\
Burns\\
Other injury
\end{tabular} \\
\cline{2-3}
& Toxicological and Iatrogenic Complications &
\begin{tabular}[t]{@{}l@{}}
Poisoning by psychiatric medications\\
Poisoning by other medications\\
Poisoning by non-medications\\
Procedure complications\\
Device complications
\end{tabular} \\
\hline

\multirow{2}{5cm}{Symptoms, Signs, and Other Conditions} &
Symptoms and Signs &
\begin{tabular}[t]{@{}l@{}}
Syncope\\
Fever of unknown origin\\
Lymph node enlargement\\
Gangrene\\
Shock\\
Nausea and vomiting\\
Abdominal pain\\
Fatigue\\
Allergy\\
Unclassified
\end{tabular} \\
\cline{2-3}
& Aftercare and Other Issues &
\begin{tabular}[t]{@{}l@{}}
Rehabilitation\\
Social and administrative problems\\
Examination and evaluation\\
Other aftercare\\
Other screening
\end{tabular} \\
\hline

\multirow{4}{5cm}{Mental; Behavioral; and Neurodevelopmental Disorders} &
Neurodevelopmental and Pediatric Disorders &
\begin{tabular}[t]{@{}l@{}}
Developmental disorders\\
Disorders diagnosed in infancy/childhood\\
Attention-deficit/conduct/disruptive \\ \hspace*{1em} behavior disorders\\
Impulse control disorders\\
Blindness
\end{tabular} \\
\cline{2-3}
& Mood; Anxiety; and Cognitive Disorders &
\begin{tabular}[t]{@{}l@{}}
Adjustment disorders\\
Anxiety disorders\\
Mood disorders\\
Personality disorders
\end{tabular} \\
\cline{2-3}
& Psychotic and Substance Use Disorders &
\begin{tabular}[t]{@{}l@{}}
Schizophrenia and other psychotic \\ \hspace*{1em} disorders\\
Alcohol-related disorders\\
Substance-related disorders
\end{tabular} \\
\cline{2-3}
& Other Conditions and Events &
\begin{tabular}[t]{@{}l@{}}
Delirium/dementia/amnesia/cognitive \\ \hspace*{1em} disorders\\
Miscellaneous mental health disorders\\
Suicide and intentional self-inflicted injury\\
Mental health screening or history
\end{tabular} \\
\hline

\multirow{7}{5cm}{Neoplasms} &
Gastrointestinal Cancers &
\begin{tabular}[t]{@{}l@{}}
Esophageal cancer\\
Stomach cancer\\
Colon cancer\\
Rectal or anal cancer\\
Pancreatic cancer\\
Gastrointestinal and peritoneal cancer
\end{tabular} \\
\cline{2-3}
& Head; Neck; and Thoracic Cancers &
\begin{tabular}[t]{@{}l@{}}
Head or neck cancer\\
Bronchus/lung cancer\\
Other respiratory cancer
\end{tabular} \\
\cline{2-3}
& Urogenital and Reproductive Cancers &
\begin{tabular}[t]{@{}l@{}}
Breast cancer\\
Uterine cancer\\
Ovary cancer\\
Cervical cancer\\
Female genital cancer\\
Prostate cancer\\
Testicular cancer\\
Male genital cancer\\
Bladder cancer\\
Kidney and renal cancer\\
Urinary organ cancer
\end{tabular} \\
\cline{2-3}
& Hematologic and Endocrine Cancers &
\begin{tabular}[t]{@{}l@{}}
Hodgkin's disease\\
Non-Hodgkin lymphoma\\
Leukemias\\
Multiple myeloma\\
Thyroid cancer
\end{tabular} \\
\cline{2-3}
& Cancers of Other Systems &
\begin{tabular}[t]{@{}l@{}}
Brain/nervous system cancer\\
Bone/connective tissue cancer\\
Skin melanoma\\
Non-epithelial cancer\\
Other primary cancer\\
Secondary malignancy\\
Liver or inflammatory bowel disease \\ \hspace*{1em} cancer
\end{tabular} \\
\cline{2-3}
& Benign Neoplasms &
\begin{tabular}[t]{@{}l@{}}
Benign uterine neoplasm\\
Other benign neoplasm
\end{tabular} \\
\cline{2-3}
& Unspecified Neoplasms and Other Conditions &
\begin{tabular}[t]{@{}l@{}}
Malignant neoplasm\\
Neoplasm unspecified\\
Maintenance chemotherapy/radiation
\end{tabular} \\
\hline

\multirow{1}{5cm}{External Causes of Morbidity (E-Codes)} &
Environmental; Mechanical; and Intentional Injuries &
\begin{tabular}[t]{@{}l@{}}
E codes: Cut or pierce\\
E codes: Drowning or submersion\\
E codes: Fall\\
E codes: Fire or burn\\
E codes: Firearm\\
E codes: Machinery\\
E codes: Motor vehicle traffic\\
E codes: Pedal cyclist\\
E codes: Pedestrian\\
E codes: Transport not motor vehicle traffic\\
E codes: Environmental causes\\
E codes: Overexertion\\
E codes: Poisoning\\
E codes: Struck by or against\\
E codes: Suffocation\\
E codes: Adverse effects of medical care\\
E codes: Adverse effects of medical drugs\\
E codes: Other specified and classifiable\\
E codes: Other specified NEC\\
E codes: Unspecified\\
E codes: Place of occurrence
\end{tabular} \\
\hline

\end{longtable}

\section{Performance and Fairness across Racial Groups under Original and Corruption Settings.}
\label{sec:appendix_disparities_by_racial_subgroups}

\begin{table}[H]
\tiny
\centering
\setlength{\tabcolsep}{3pt}
\renewcommand{\arraystretch}{1.2}
\begin{tabular}{cc*{12}{c}}
\toprule
\midrule
\multirow{5}{*}{\textbf{Model}} & \multirow{5}{*}{\textbf{Type}} 
  & \multicolumn{12}{c}{\textbf{Racial Groups (Set 1)}} \\
\cmidrule(lr){3-14}
  & & \multicolumn{4}{c}{WHITE} & \multicolumn{4}{c}{BLACK} & \multicolumn{4}{c}{HISPANIC\_LATINO} \\
\cmidrule(lr){3-6} \cmidrule(lr){7-10} \cmidrule(lr){11-14}
  & & \multicolumn{2}{c}{Recall@10} & \multicolumn{2}{c}{Precision@10} 
    & \multicolumn{2}{c}{Recall@10} & \multicolumn{2}{c}{Precision@10}
    & \multicolumn{2}{c}{Recall@10} & \multicolumn{2}{c}{Precision@10} \\
\cmidrule(lr){3-4}\cmidrule(lr){5-6}\cmidrule(lr){7-8}\cmidrule(lr){9-10}
\cmidrule(lr){11-12}\cmidrule(lr){13-14}
  & & Mean & 95CI RW & Mean & 95CI RW
    & Mean & 95CI RW & Mean & 95CI RW
    & Mean & 95CI RW & Mean & 95CI RW \\
\hline
\midrule
\multirow{5}{*}{\texttt{Gemini-2.0-Flash}}
  & Original               & 0.4698 & 0.0205 & 0.3939 & 0.0245 & 0.5038 & 0.0436 & 0.3973 & 0.0546 & 0.5172 & 0.0850 & 0.3725 & 0.0972 \\
  & Lab-Value Erasure      & 0.4700 & 0.0212 & 0.3929 & 0.0244 & 0.4992 & 0.0453 & 0.3956 & 0.0563 & 0.5134 & 0.0849 & 0.3719 & 0.1038 \\
  & Prior-note Duplication & 0.4714 & 0.0208 & 0.3944 & 0.0240 & 0.5019 & 0.0443 & 0.3984 & 0.0553 & 0.5242 & 0.0805 & 0.3786 & 0.0992 \\
  & OCR Jittering          & 0.4729 & 0.0197 & 0.3963 & 0.0240 & 0.5034 & 0.0476 & 0.3988 & 0.0574 & 0.5176 & 0.0790 & 0.3777 & 0.1015 \\
  & Homophone Substitution & 0.4716 & 0.0204 & 0.3964 & 0.0244 & 0.5015 & 0.0482 & 0.3957 & 0.0557 & 0.5176 & 0.0830 & 0.3777 & 0.1015 \\
\midrule
\multirow{5}{*}{\texttt{GPT-4o-mini}}
  & Original               & 0.3552 & 0.0256 & 0.2976 & 0.0264 & 0.3790 & 0.0554 & 0.2968 & 0.0604 & 0.3917 & 0.0999 & 0.2801 & 0.1071 \\
  & Lab-Value Erasure      & 0.3546 & 0.0232 & 0.2962 & 0.0249 & 0.3800 & 0.0549 & 0.2989 & 0.0607 & 0.3982 & 0.0965 & 0.2839 & 0.1045 \\
  & Prior-note Duplication & 0.3534 & 0.0245 & 0.2963 & 0.0268 & 0.3730 & 0.0587 & 0.2921 & 0.0604 & 0.3944 & 0.1029 & 0.2821 & 0.1082 \\
  & OCR Jittering          & 0.3535 & 0.0248 & 0.2960 & 0.0269 & 0.3769 & 0.0572 & 0.2971 & 0.0600 & 0.4013 & 0.1033 & 0.2858 & 0.1139 \\
  & Homophone Substitution & 0.3559 & 0.0236 & 0.2979 & 0.0268 & 0.3797 & 0.0570 & 0.2958 & 0.0587 & 0.3923 & 0.1027 & 0.2778 & 0.1061 \\
\bottomrule
\end{tabular}
\caption{Performance under Corruption Settings across Racial Subgroups (Set 1). RW = relative width of 95\% CI.}
\label{tab:overall_performance_race_set1}
\end{table}

\begin{table}[H]
\tiny
\centering
\setlength{\tabcolsep}{3pt}
\renewcommand{\arraystretch}{1.2}
\begin{tabular}{cc*{12}{c}}
\toprule
\midrule
\multirow{5}{*}{\textbf{Model}} & \multirow{5}{*}{\textbf{Type}} 
  & \multicolumn{12}{c}{\textbf{Racial Groups (Set 2)}} \\
\cmidrule(lr){3-14}
  & & \multicolumn{4}{c}{OTHER} & \multicolumn{4}{c}{ASIAN} & \multicolumn{4}{c}{UNKNOWN} \\
\cmidrule(lr){3-6} \cmidrule(lr){7-10} \cmidrule(lr){11-14}
  & & \multicolumn{2}{c}{Recall@10} & \multicolumn{2}{c}{Precision@10} 
    & \multicolumn{2}{c}{Recall@10} & \multicolumn{2}{c}{Precision@10}
    & \multicolumn{2}{c}{Recall@10} & \multicolumn{2}{c}{Precision@10} \\
\cmidrule(lr){3-4}\cmidrule(lr){5-6}\cmidrule(lr){7-8}\cmidrule(lr){9-10}
\cmidrule(lr){11-12}\cmidrule(lr){13-14}
  & & Mean & 95CI RW & Mean & 95CI RW
    & Mean & 95CI RW & Mean & 95CI RW
    & Mean & 95CI RW & Mean & 95CI RW \\
\midrule
\multirow{5}{*}{\texttt{Gemini-2.0-Flash}}
  & Original               & 0.4819 & 0.1005 & 0.3796 & 0.1142 & 0.5093 & 0.1004 & 0.3714 & 0.1245 & 0.4758 & 0.1114 & 0.3775 & 0.1343 \\
  & Lab-Value Erasure      & 0.4831 & 0.1023 & 0.3776 & 0.1148 & 0.5067 & 0.1070 & 0.3686 & 0.1311 & 0.4802 & 0.1081 & 0.3826 & 0.1313 \\
  & Prior-note Duplication & 0.4824 & 0.1055 & 0.3786 & 0.1135 & 0.5089 & 0.0998 & 0.3700 & 0.1318 & 0.4744 & 0.1123 & 0.3754 & 0.1302 \\
  & OCR Jittering          & 0.4746 & 0.1053 & 0.3696 & 0.1162 & 0.5091 & 0.1047 & 0.3743 & 0.1292 & 0.4834 & 0.1187 & 0.3834 & 0.1357 \\
  & Homophone Substitution & 0.4918 & 0.1024 & 0.3865 & 0.1121 & 0.5207 & 0.1006 & 0.3751 & 0.1267 & 0.4795 & 0.1145 & 0.3787 & 0.1387 \\
\midrule
\multirow{5}{*}{\texttt{GPT-4o-mini}}
  & Original               & 0.3808 & 0.1097 & 0.2933 & 0.1193 & 0.4058 & 0.1318 & 0.2933 & 0.1435 & 0.3637 & 0.1330 & 0.2900 & 0.1433 \\
  & Lab-Value Erasure      & 0.3828 & 0.1199 & 0.2954 & 0.1254 & 0.4029 & 0.1250 & 0.2925 & 0.1410 & 0.3647 & 0.1377 & 0.2884 & 0.1498 \\
  & Prior-note Duplication & 0.3798 & 0.1242 & 0.2928 & 0.1238 & 0.3984 & 0.1293 & 0.2836 & 0.1338 & 0.3666 & 0.1259 & 0.2928 & 0.1413 \\
  & OCR Jittering          & 0.3913 & 0.1212 & 0.2985 & 0.1151 & 0.4044 & 0.1189 & 0.2933 & 0.1392 & 0.3673 & 0.1261 & 0.2969 & 0.1446 \\
  & Homophone Substitution & 0.3736 & 0.1236 & 0.2864 & 0.1221 & 0.4099 & 0.1313 & 0.2977 & 0.1428 & 0.3675 & 0.1295 & 0.2940 & 0.1462 \\
\bottomrule
\end{tabular}
\caption{Performance under Corruption Settings across Racial Subgroups (Set 2). RW = relative width of 95\% CI.}
\label{tab:overall_performance_race_set2}
\end{table}

\begin{table}[H]
\tiny
\centering
\setlength{\tabcolsep}{3pt}
\renewcommand{\arraystretch}{1.2}
\begin{tabular}{cc*{12}{c}}
\toprule
\midrule
\multirow{5}{*}{\textbf{Model}} & \multirow{5}{*}{\textbf{Type}} 
  & \multicolumn{12}{c}{\textbf{Racial Groups (Set 1)}} \\
\cmidrule(lr){3-14}
  & & \multicolumn{4}{c}{WHITE} & \multicolumn{4}{c}{BLACK} & \multicolumn{4}{c}{HISPANIC\_LATINO} \\
\cmidrule(lr){3-6} \cmidrule(lr){7-10} \cmidrule(lr){11-14}
  & & \multicolumn{2}{c}{TPR} & \multicolumn{2}{c}{FPR} 
    & \multicolumn{2}{c}{TPR} & \multicolumn{2}{c}{FPR}
    & \multicolumn{2}{c}{TPR} & \multicolumn{2}{c}{FPR} \\
\cmidrule(lr){3-4}\cmidrule(lr){5-6}\cmidrule(lr){7-8}\cmidrule(lr){9-10}
\cmidrule(lr){11-12}\cmidrule(lr){13-14}
  & & Mean & 95CI RW & Mean & 95CI RW
    & Mean & 95CI RW & Mean & 95CI RW
    & Mean & 95CI RW & Mean & 95CI RW \\
\hline
\midrule
\multirow{5}{*}{\texttt{Gemini-2.0-Flash}}
  & Original               & 0.0777 & 0.0284 & 0.0593 & 0.0384 & 0.0809 & 0.0641 & 0.0620 & 0.0794 & 0.0745 & 0.1111 & 0.0594 & 0.1392 \\
  & Lab-Value Erasure      & 0.0772 & 0.0286 & 0.0590 & 0.0385 & 0.0802 & 0.0666 & 0.0618 & 0.0837 & 0.0750 & 0.1178 & 0.0586 & 0.1433 \\
  & Prior-note Duplication & 0.0773 & 0.0277 & 0.0581 & 0.0378 & 0.0804 & 0.0650 & 0.0603 & 0.0829 & 0.0752 & 0.1124 & 0.0568 & 0.1337 \\
  & OCR Jittering          & 0.0782 & 0.0285 & 0.0602 & 0.0369 & 0.0816 & 0.0643 & 0.0630 & 0.0802 & 0.0752 & 0.1183 & 0.0596 & 0.1437 \\
  & Homophone Substitution & 0.0779 & 0.0286 & 0.0593 & 0.0360 & 0.0804 & 0.0648 & 0.0629 & 0.0803 & 0.0751 & 0.1170 & 0.0602 & 0.1385 \\
\midrule
\multirow{5}{*}{\texttt{GPT-4o-mini}}
  & Original               & 0.0488 & 0.0286 & 0.0424 & 0.0349 & 0.0501 & 0.0638 & 0.0427 & 0.0779 & 0.0474 & 0.1112 & 0.0411 & 0.1447 \\
  & Lab-Value Erasure      & 0.0485 & 0.0267 & 0.0425 & 0.0347 & 0.0505 & 0.0669 & 0.0440 & 0.0764 & 0.0482 & 0.1103 & 0.0414 & 0.1364 \\
  & Prior-note Duplication & 0.0485 & 0.0288 & 0.0425 & 0.0350 & 0.0497 & 0.0638 & 0.0436 & 0.0772 & 0.0473 & 0.1210 & 0.0426 & 0.1317 \\
  & OCR Jittering          & 0.0486 & 0.0306 & 0.0427 & 0.0342 & 0.0501 & 0.0648 & 0.0430 & 0.0813 & 0.0477 & 0.1222 & 0.0409 & 0.1467 \\
  & Homophone Substitution & 0.0487 & 0.0290 & 0.0426 & 0.0336 & 0.0501 & 0.0610 & 0.0438 & 0.0745 & 0.0468 & 0.1174 & 0.0414 & 0.1410 \\
\bottomrule
\end{tabular}
\caption{Fairness under Corruption Settings across Racial Subgroups (Set 1). RW = relative width of 95\% CI.}
\label{tab:overall_fairness_race_set1}
\end{table}

\begin{table}[H]
\tiny
\centering
\setlength{\tabcolsep}{3pt}
\renewcommand{\arraystretch}{1.2}
\begin{tabular}{cc*{12}{c}}
\toprule
\midrule
\multirow{5}{*}{\textbf{Model}} & \multirow{5}{*}{\textbf{Type}} 
  & \multicolumn{12}{c}{\textbf{Racial Groups (Set 2)}} \\
\cmidrule(lr){3-14}
  & & \multicolumn{4}{c}{OTHER} & \multicolumn{4}{c}{ASIAN} & \multicolumn{4}{c}{UNKNOWN} \\
\cmidrule(lr){3-6} \cmidrule(lr){7-10} \cmidrule(lr){11-14}
  & & \multicolumn{2}{c}{TPR} & \multicolumn{2}{c}{FPR} 
    & \multicolumn{2}{c}{TPR} & \multicolumn{2}{c}{FPR}
    & \multicolumn{2}{c}{TPR} & \multicolumn{2}{c}{FPR} \\
\cmidrule(lr){3-4}\cmidrule(lr){5-6}\cmidrule(lr){7-8}\cmidrule(lr){9-10}
\cmidrule(lr){11-12}\cmidrule(lr){13-14}
  & & Mean & 95CI RW & Mean & 95CI RW
    & Mean & 95CI RW & Mean & 95CI RW
    & Mean & 95CI RW & Mean & 95CI RW \\
\hline
\midrule
\multirow{5}{*}{\texttt{Gemini-2.0-Flash}}
  & Original               & 0.0721 & 0.1307 & 0.0559 & 0.1788 & 0.0732 & 0.1461 & 0.0563 & 0.1691 & 0.0714 & 0.1474 & 0.0460 & 0.1798 \\
  & Lab-Value Erasure      & 0.0723 & 0.1339 & 0.0547 & 0.1783 & 0.0728 & 0.1542 & 0.0559 & 0.1784 & 0.0719 & 0.1463 & 0.0451 & 0.1985 \\
  & Prior-note Duplication & 0.0726 & 0.1275 & 0.0542 & 0.1798 & 0.0722 & 0.1555 & 0.0542 & 0.1690 & 0.0711 & 0.1426 & 0.0466 & 0.1753 \\
  & OCR Jittering          & 0.0701 & 0.1380 & 0.0576 & 0.1809 & 0.0731 & 0.1500 & 0.0553 & 0.1854 & 0.0733 & 0.1476 & 0.0468 & 0.1978 \\
  & Homophone Substitution & 0.0740 & 0.1252 & 0.0545 & 0.1897 & 0.0735 & 0.1455 & 0.0572 & 0.1821 & 0.0701 & 0.1544 & 0.0445 & 0.1965 \\
\midrule
\multirow{5}{*}{\texttt{GPT-4o-mini}}
  & Original               & 0.0475 & 0.1324 & 0.0415 & 0.1593 & 0.0480 & 0.1530 & 0.0417 & 0.1675 & 0.0476 & 0.1542 & 0.0352 & 0.1918 \\
  & Lab-Value Erasure      & 0.0480 & 0.1314 & 0.0409 & 0.1683 & 0.0492 & 0.1548 & 0.0420 & 0.1725 & 0.0471 & 0.1638 & 0.0366 & 0.1785 \\
  & Prior-note Duplication & 0.0475 & 0.1310 & 0.0411 & 0.1577 & 0.0464 & 0.1542 & 0.0416 & 0.1634 & 0.0485 & 0.1571 & 0.0368 & 0.1883 \\
  & OCR Jittering          & 0.0487 & 0.1261 & 0.0413 & 0.1649 & 0.0486 & 0.1585 & 0.0430 & 0.1685 & 0.0484 & 0.1643 & 0.0372 & 0.1977 \\
  & Homophone Substitution & 0.0457 & 0.1304 & 0.0414 & 0.1735 & 0.0494 & 0.1595 & 0.0414 & 0.1753 & 0.0478 & 0.1626 & 0.0384 & 0.1848 \\
\bottomrule
\end{tabular}
\caption{Fairness under Corruption Settings across Racial Subgroups (Set 2). RW = relative width of 95\% CI.}
\label{tab:overall_fairness_race_set2}
\end{table}

\section{Performance and Fairness across Age Groups under Original and Corruption Settings.}
\label{sec:appendix_disparities_by_age_subgroups}

\begin{table}[H]
\tiny
\centering
\setlength{\tabcolsep}{3pt}
\renewcommand{\arraystretch}{1.2}
\begin{tabular}{cc*{12}{c}}
\toprule
\midrule
\multirow{5}{*}{\textbf{Model}} & \multirow{5}{*}{\textbf{Type}} 
  & \multicolumn{12}{c}{\textbf{Age Groups}} \\
\cmidrule(lr){3-14}
  & & \multicolumn{4}{c}{18–40} & \multicolumn{4}{c}{41–60} & \multicolumn{4}{c}{61+} \\
\cmidrule(lr){3-6} \cmidrule(lr){7-10} \cmidrule(lr){11-14}
  & & \multicolumn{2}{c}{Recall@10} & \multicolumn{2}{c}{Precision@10} 
    & \multicolumn{2}{c}{Recall@10} & \multicolumn{2}{c}{Precision@10}
    & \multicolumn{2}{c}{Recall@10} & \multicolumn{2}{c}{Precision@10} \\
\cmidrule(lr){3-4}\cmidrule(lr){5-6}\cmidrule(lr){7-8}\cmidrule(lr){9-10}
\cmidrule(lr){11-12}\cmidrule(lr){13-14}
  & & Mean & 95CI RW & Mean & 95CI RW
    & Mean & 95CI RW & Mean & 95CI RW
    & Mean & 95CI RW & Mean & 95CI RW \\
\hline
\midrule
\multirow{5}{*}{\texttt{Gemini-2.0-Flash}}
  & Original               & 0.5383 & 0.0523 & 0.2958 & 0.0684 & 0.4776 & 0.0325 & 0.3698 & 0.0404 & 0.4648 & 0.0212 & 0.4273 & 0.0246 \\
  & Lab-Value Erasure      & 0.5398 & 0.0541 & 0.2941 & 0.0694 & 0.4770 & 0.0323 & 0.3695 & 0.0393 & 0.4636 & 0.0211 & 0.4260 & 0.0243 \\
  & Prior-note Duplication & 0.5355 & 0.0552 & 0.2948 & 0.0683 & 0.4798 & 0.0316 & 0.3714 & 0.0406 & 0.4664 & 0.0219 & 0.4279 & 0.0250 \\
  & OCR Jittering          & 0.5359 & 0.0532 & 0.2940 & 0.0685 & 0.4785 & 0.0329 & 0.3700 & 0.0395 & 0.4684 & 0.0217 & 0.4314 & 0.0246 \\
  & Homophone Substitution & 0.5333 & 0.0540 & 0.2919 & 0.0677 & 0.4791 & 0.0334 & 0.3710 & 0.0401 & 0.4681 & 0.0215 & 0.4315 & 0.0246 \\
\midrule
\multirow{5}{*}{\texttt{GPT-4o-mini}}
  & Original               & 0.4235 & 0.0681 & 0.2237 & 0.0715 & 0.3556 & 0.0408 & 0.2747 & 0.0436 & 0.3519 & 0.0243 & 0.3254 & 0.0283 \\
  & Lab-Value Erasure      & 0.4290 & 0.0661 & 0.2267 & 0.0699 & 0.3547 & 0.0416 & 0.2733 & 0.0443 & 0.3514 & 0.0249 & 0.3247 & 0.0282 \\
  & Prior-note Duplication & 0.4249 & 0.0651 & 0.2255 & 0.0679 & 0.3524 & 0.0406 & 0.2725 & 0.0447 & 0.3495 & 0.0246 & 0.3231 & 0.0287 \\
  & OCR Jittering          & 0.4238 & 0.0644 & 0.2240 & 0.0694 & 0.3565 & 0.0402 & 0.2745 & 0.0439 & 0.3503 & 0.0241 & 0.3248 & 0.0275 \\
  & Homophone Substitution & 0.4283 & 0.0662 & 0.2223 & 0.0667 & 0.3566 & 0.0422 & 0.2756 & 0.0436 & 0.3514 & 0.0232 & 0.3253 & 0.0271 \\
\bottomrule
\end{tabular}
\caption{Performance under Corruption Settings across Age Subgroups. RW = relative width of 95\% CI.}
\label{tab:overall_performance_age}
\end{table}

\begin{table}[H]
\tiny
\centering
\setlength{\tabcolsep}{3pt}
\renewcommand{\arraystretch}{1.2}
\begin{tabular}{cc*{12}{c}}
\toprule
\midrule
\multirow{5}{*}{\textbf{Model}} & \multirow{5}{*}{\textbf{Type}} 
  & \multicolumn{12}{c}{\textbf{Age Groups}} \\
\cmidrule(lr){3-14}
  & & \multicolumn{4}{c}{18--40} & \multicolumn{4}{c}{41--60} & \multicolumn{4}{c}{61+} \\
\cmidrule(lr){3-6} \cmidrule(lr){7-10} \cmidrule(lr){11-14}
  & & \multicolumn{2}{c}{TPR} & \multicolumn{2}{c}{FPR} 
    & \multicolumn{2}{c}{TPR} & \multicolumn{2}{c}{FPR}
    & \multicolumn{2}{c}{TPR} & \multicolumn{2}{c}{FPR} \\
\cmidrule(lr){3-4}\cmidrule(lr){5-6}\cmidrule(lr){7-8}\cmidrule(lr){9-10}
\cmidrule(lr){11-12}\cmidrule(lr){13-14}
  & & Mean & 95CI RW & Mean & 95CI RW
    & Mean & 95CI RW & Mean & 95CI RW
    & Mean & 95CI RW & Mean & 95CI RW \\
\hline
\midrule
\multirow{5}{*}{\texttt{Gemini-2.0-Flash}}
  & Original               & 0.0567 & 0.0755 & 0.0555 & 0.0849 & 0.0741 & 0.0457 & 0.0592 & 0.0555 & 0.0845 & 0.0289 & 0.0599 & 0.0438 \\
  & Lab-Value Erasure      & 0.0566 & 0.0771 & 0.0544 & 0.0863 & 0.0736 & 0.0442 & 0.0585 & 0.0560 & 0.0840 & 0.0299 & 0.0600 & 0.0416 \\
  & Prior-note Duplication & 0.0561 & 0.0763 & 0.0543 & 0.0832 & 0.0735 & 0.0460 & 0.0571 & 0.0549 & 0.0843 & 0.0293 & 0.0589 & 0.0430 \\
  & OCR Jittering          & 0.0560 & 0.0756 & 0.0556 & 0.0843 & 0.0741 & 0.0454 & 0.0601 & 0.0568 & 0.0855 & 0.0293 & 0.0611 & 0.0431 \\
  & Homophone Substitution & 0.0558 & 0.0766 & 0.0551 & 0.0887 & 0.0737 & 0.0458 & 0.0592 & 0.0578 & 0.0851 & 0.0291 & 0.0603 & 0.0430 \\
\midrule
\multirow{5}{*}{\texttt{GPT-4o-mini}}
  & Original               & 0.0365 & 0.0765 & 0.0403 & 0.0806 & 0.0456 & 0.0480 & 0.0429 & 0.0562 & 0.0535 & 0.0307 & 0.0423 & 0.0385 \\
  & Lab-Value Erasure      & 0.0365 & 0.0788 & 0.0407 & 0.0832 & 0.0457 & 0.0465 & 0.0434 & 0.0536 & 0.0534 & 0.0301 & 0.0424 & 0.0397 \\
  & Prior-note Duplication & 0.0366 & 0.0774 & 0.0407 & 0.0822 & 0.0452 & 0.0477 & 0.0435 & 0.0523 & 0.0533 & 0.0315 & 0.0422 & 0.0388 \\
  & OCR Jittering          & 0.0365 & 0.0787 & 0.0415 & 0.0823 & 0.0454 & 0.0475 & 0.0434 & 0.0535 & 0.0537 & 0.0298 & 0.0422 & 0.0379 \\
  & Homophone Substitution & 0.0359 & 0.0756 & 0.0416 & 0.0816 & 0.0458 & 0.0469 & 0.0429 & 0.0566 & 0.0535 & 0.0311 & 0.0426 & 0.0373 \\
\bottomrule
\end{tabular}
\caption{Fairness under corruption settings across age subgroups. RW = relative width of 95\% CI.}
\label{tab:overall_fairness_age}
\end{table}

\section{Performance and Fairness across Sex Groups under Original and Corruption Settings.}
\label{sec:appendix_disparities_by_sex_subgroups}

\begin{table}[H]
\tiny
\centering
\setlength{\tabcolsep}{3pt}
\renewcommand{\arraystretch}{1.2}
\begin{tabular}{cc*{8}{c}}
\toprule
\midrule
\multirow{5}{*}{\textbf{Model}} & \multirow{5}{*}{\textbf{Type}} 
  & \multicolumn{8}{c}{\textbf{Sex Groups}} \\
\cmidrule(lr){3-10}
  & & \multicolumn{4}{c}{Female} & \multicolumn{4}{c}{Male} \\
\cmidrule(lr){3-6} \cmidrule(lr){7-10}
  & & \multicolumn{2}{c}{Recall@10} & \multicolumn{2}{c}{Precision@10} 
    & \multicolumn{2}{c}{Recall@10} & \multicolumn{2}{c}{Precision@10} \\
\cmidrule(lr){3-4}\cmidrule(lr){5-6}\cmidrule(lr){7-8}\cmidrule(lr){9-10}
  & & Mean & 95CI RW & Mean & 95CI RW
    & Mean & 95CI RW & Mean & 95CI RW \\
\hline
\midrule
\multirow{5}{*}{\texttt{Gemini-2.0-Flash}}
  & Original               & 0.4763 & 0.0251 & 0.3855 & 0.0303 & 0.4812 & 0.0244 & 0.3982 & 0.0291 \\
  & Lab-Value Erasure      & 0.4758 & 0.0253 & 0.3842 & 0.0319 & 0.4805 & 0.0248 & 0.3975 & 0.0294 \\
  & Prior-note Duplication & 0.4786 & 0.0257 & 0.3860 & 0.0302 & 0.4813 & 0.0241 & 0.3990 & 0.0287 \\
  & OCR Jittering          & 0.4806 & 0.0242 & 0.3886 & 0.0305 & 0.4812 & 0.0246 & 0.3993 & 0.0294 \\
  & Homophone Substitution & 0.4786 & 0.0247 & 0.3877 & 0.0308 & 0.4824 & 0.0242 & 0.4004 & 0.0289 \\
\midrule
\multirow{5}{*}{\texttt{GPT-4o-mini}}
  & Original               & 0.3633 & 0.0302 & 0.2903 & 0.0323 & 0.3626 & 0.0294 & 0.3020 & 0.0325 \\
  & Lab-Value Erasure      & 0.3639 & 0.0300 & 0.2906 & 0.0327 & 0.3620 & 0.0293 & 0.3006 & 0.0324 \\
  & Prior-note Duplication & 0.3617 & 0.0299 & 0.2901 & 0.0320 & 0.3598 & 0.0284 & 0.2987 & 0.0321 \\
  & OCR Jittering          & 0.3628 & 0.0293 & 0.2902 & 0.0315 & 0.3618 & 0.0295 & 0.3013 & 0.0326 \\
  & Homophone Substitution & 0.3634 & 0.0308 & 0.2902 & 0.0319 & 0.3638 & 0.0297 & 0.3021 & 0.0330 \\
\bottomrule
\end{tabular}
\caption{Performance under Corruption Settings across Sex Subgroups. RW = relative width of 95\% CI.}
\label{tab:overall_performance_sex}
\end{table}

\begin{table}[H]
\tiny
\centering
\setlength{\tabcolsep}{3pt}
\renewcommand{\arraystretch}{1.2}
\begin{tabular}{cc*{8}{c}}
\toprule
\midrule
\multirow{5}{*}{\textbf{Model}} & \multirow{5}{*}{\textbf{Type}} 
  & \multicolumn{8}{c}{\textbf{Sex Groups}} \\
\cmidrule(lr){3-10}
  & & \multicolumn{4}{c}{Female} & \multicolumn{4}{c}{Male} \\
\cmidrule(lr){3-6} \cmidrule(lr){7-10}
  & & \multicolumn{2}{c}{TPR} & \multicolumn{2}{c}{FPR} 
    & \multicolumn{2}{c}{TPR} & \multicolumn{2}{c}{FPR} \\
\cmidrule(lr){3-4}\cmidrule(lr){5-6}\cmidrule(lr){7-8}\cmidrule(lr){9-10}
  & & Mean & 95CI RW & Mean & 95CI RW
    & Mean & 95CI RW & Mean & 95CI RW \\
\midrule
\multirow{5}{*}{\texttt{Gemini-2.0-Flash}}
  & Original               & 0.0762 & 0.0359 & 0.0594 & 0.0456 & 0.0789 & 0.0328 & 0.0588 & 0.0447 \\
  & Lab-Value Erasure      & 0.0758 & 0.0366 & 0.0590 & 0.0461 & 0.0785 & 0.0328 & 0.0584 & 0.0446 \\
  & Prior-note Duplication & 0.0756 & 0.0350 & 0.0579 & 0.0443 & 0.0788 & 0.0332 & 0.0576 & 0.0434 \\
  & OCR Jittering          & 0.0769 & 0.0348 & 0.0606 & 0.0464 & 0.0791 & 0.0350 & 0.0594 & 0.0449 \\
  & Homophone Substitution & 0.0763 & 0.0358 & 0.0597 & 0.0457 & 0.0790 & 0.0330 & 0.0587 & 0.0441 \\
\midrule
\multirow{5}{*}{\texttt{GPT-4o-mini}}
  & Original               & 0.0480 & 0.0343 & 0.0421 & 0.0403 & 0.0496 & 0.0344 & 0.0422 & 0.0443 \\
  & Lab-Value Erasure      & 0.0479 & 0.0347 & 0.0423 & 0.0411 & 0.0495 & 0.0347 & 0.0426 & 0.0435 \\
  & Prior-note Duplication & 0.0480 & 0.0359 & 0.0422 & 0.0417 & 0.0492 & 0.0354 & 0.0426 & 0.0414 \\
  & OCR Jittering          & 0.0479 & 0.0349 & 0.0422 & 0.0410 & 0.0497 & 0.0352 & 0.0427 & 0.0433 \\
  & Homophone Substitution & 0.0480 & 0.0346 & 0.0426 & 0.0402 & 0.0495 & 0.0348 & 0.0425 & 0.0419 \\
\bottomrule
\end{tabular}
\caption{Fairness under Corruption Settings across Sex Subgroups. RW = relative width of 95\% CI.}
\label{tab:overall_fairness_sex}
\end{table}

\end{document}